\documentclass[12pt]{article}
\usepackage{times}
\usepackage{epsfig}
\usepackage{amssymb}
\usepackage{xcolor}
\usepackage{amsmath,comment}
\usepackage{algorithm}
\usepackage[noend]{algpseudocode}

\DeclareMathOperator*{\argmax}{arg\,max}
\usepackage{caption}
\usepackage{subcaption}
\usepackage{makecell}
\usepackage{mathtools}
\usepackage{bbm}
\DeclarePairedDelimiter{\ceil}{\lceil}{\rceil}
\DeclarePairedDelimiter{\floor}{\lfloor}{\rfloor}
\newtheorem{property}{Property}
\usepackage{url}
\usepackage{tikz}

\begin{document}
\title{Revealing Perceptible Backdoors in DNNs, Without the Training Set, via the Maximum Achievable Misclassification Fraction Statistic\thanks{This research supported by an AFOSR DDDAS grant and a Cisco Systems URP gift.}}

\author{Zhen Xiang, David J. Miller, Hang Wang, and George Kesidis\\
School of EECS, The Pennsylvania State University\\
University Park, PA 16802 USA\\
\{zux49,djm25,hzw81,gik2\}@psu.edu
	}
\maketitle 

\begin{abstract}
Recently, a backdoor data poisoning attack was proposed, which adds mislabeled examples to the training set, with an embedded backdoor pattern, aiming to have the classifier learn to classify to a target class whenever the backdoor pattern is present in a test sample. Here, we address {\it post-training} detection of innocuous {\it perceptible} backdoors in DNN image classifiers, wherein the defender does {\it not} have access to the poisoned training set, but only to the trained classifier, as well as unpoisoned examples. This problem is challenging because without the poisoned training set, we have no hint about the actual backdoor pattern used during training. This post-training scenario is also of great import because in many practical contexts the DNN user did not train the DNN and does not have access to the training data.  We identify two important properties of perceptible backdoor patterns -- spatial invariance and robustness -- based upon which we propose a novel detector using the {\it maximum achievable misclassification fraction} (MAMF) statistic. We detect whether the trained DNN has been backdoor-attacked and infer the source and target classes. Our detector outperforms other existing detectors and, coupled with an {\it imperceptible} backdoor detector, helps achieve post-training detection of all evasive backdoors.
\end{abstract}

\section{Introduction}\label{sec:introduction}
Deep neural network (DNN) classifiers have achieved state-of-the-art pattern recognition performance in many research areas such as speech recognition \cite{Speech_ref1}, bioinformatics \cite{Bio_ref1}, and computer vision \cite{CV_ref1}\cite{CV_ref2}. However, they have also been shown to be vulnerable to adversarial attacks \cite{Szegedy_seminal}. This has inspired adversarial learning research, including work in devising both formidable attacks as well as defenses against same \cite{Review}. Test-time evasion (TTE) is a prominent type of adversarial attack aiming to induce misclassifications during operation/test-time by modifying test samples in a human-imperceptible (or machine-evasive) fashion, e.g. \cite{FGSM}\cite{JSMA}\cite{DeepFool}\cite{CW}\cite{EAD}. Another type of attack, data poisoning (DP), inserts malicious samples into the training set, {\it often} to degrade the classifier's accuracy \cite{Tygar11}\cite{Xiao15}.

Recently, a new form of {\it backdoor} DP attack, usually against DNN image classifiers, was proposed, aiming to ``sneak'' a backdoor mapping into the learning of the DNN, while {\it not} degrading the performance in accurately classifying clean test samples \cite{Targeted}\cite{Trojan}\cite{BadNet}. A learned backdoor can be easily achieved by inserting a relatively small number of poisoned images into the training set\footnote{The attacker’s poisoning capability is facilitated by the need in practice to obtain ``big data'' suitable for accurately training a DNN for a given domain -- to do so, one may need to seek data from as many sources as possible (some of which could be attackers).}. Backdoor training images are crafted by embedding the same backdoor pattern into legitimate (clean) images from one or more source classes, and (mis)labeling to a target class (selected by the attacker). The backdoor pattern, principally designed to be evasive to possible occasional human inspection of the training set, could be: 1) a human-imperceptible, additive perturbation applied to clean images (dubbed here an {\it imperceptible} backdoor pattern) \cite{Haoti}\cite{SS}\cite{CI}\cite{Post-TNNLS}; 2) a seemingly innocuous object in a scene (dubbed here a {\it perceptible} backdoor pattern), e.g. a bird flying in the sky or glasses on a face \cite{Targeted}\cite{BadNet}\cite{Tabor}. The DNN trained on the poisoned training set will still correctly classify clean test images with high accuracy (because the number of poisoned samples is relatively small); hence, validation set accuracy degradation, e.g. \cite{Roni}, {\it cannot} be reliably used as a basis for detecting backdoors. However, when faced with a test image containing the same backdoor pattern used in training, the DNN is likely to misclassify to the target class prescribed by the attacker.

Defenses against backdoor attacks can be deployed before/during training, post-training, or potentially in-flight (i.e. at operational/test-time).\footnote{There are currently no published approaches for in-flight detection, to our knowledge.} In the before/during-training scenario, the defender has access to the (possibly poisoned) training set and the trained classifier \cite{SS}\cite{AC}\cite{CI}. The defender's goals are to detect whether the training set has been poisoned and, if so, to identify and remove the embedded backdoor training images before (re)training. In the post-training scenario considered here, however, the defender has access to the trained DNN but {\it not} to the training set used for its learning \cite{FP}\cite{NC}\cite{Tabor}\cite{Post-TNNLS}. This post-training scenario is of strong interest because there are many pure {\it consumers} of machine learning systems. For example, critical infrastructure is often based on {\it legacy} system classifiers. In such a scenario, it is very possible the original training data used to build the classifier is unavailable. Likewise, an app may be used on millions of cell phones. The app user will not have access to the training set on which the app's classifier was learned. Still, the user would like to know whether the app's classifier has been backdoor-poisoned. The user is assumed to possess a clean labeled dataset with examples from each of the classes in the domain on which, e.g., it can evaluate the performance of the DNN. The clean dataset is also assumed to be relatively small (this is reasonable, as the user may not have access to either substantial training data nor to computational resources necessary for DNN training); hence, while useful for building a backdoor detector, it is not sufficient for training a surrogate DNN (i.e. one without a backdoor present). The defender's fundamental goal in the post-training scenario is to detect whether the DNN has been backdoor-poisoned or not. If a detection is made, the source class(es) and the target class should also be inferred, so that decisions made to classes {\it not} involved in the attack could still possibly be trusted when there are no replacement DNNs available.

In this paper, we focus on {\it post-training} detection of backdoor attacks with {\it perceptible} backdoor patterns, which is a very challenging problem with no well-founded method proposed yet. Elsewhere, post-training detection of {\it imperceptible} backdoors is addressed \cite{Post-TNNLS}. Together, our work and the work for the imperceptible case  \cite{Post-TNNLS}``cover'' all cases of interests, providing a complete solution to post-training detection of evasive backdoors. Our main contributions are as follows:

1) We propose a novel approach for detecting backdoor attacks with perceptible backdoor patterns, post-training. Our detection is based on the {\it maximum achievable misclassification fraction} (MAMF) statistic, obtained by essentially reverse-engineering the putative backdoor pattern for each (source, target) class pair. Our detection inference uses an easily chosen threshold. No unrealistic assumptions on the shape, spatial location or object type of the backdoor pattern are made, unlike \cite{Tabor}.

2) We identify and experimentally analyze two important properties -- spatial invariance and robustness -- of perceptible backdoor patterns that are applicable to general attack instances. These properties, which have neither been identified nor exploited in prior works, are the basis of our detector and may also potentially inspire other defenses.

3) We perform substantial experimental evaluation and show the strong capability of our detector, both in general and compared against the two existing methods \cite{NC}\cite{FP} for the post-training detection problem we address. Our experiments involve five datasets of practical interest, five commonly used DNN structures, nine backdoor patterns of different types and two compared detection methods.

The rest of the paper is organized as follows. In Section \ref{sec:backdoor}, we provide a thorough and unified review of all major aspects of backdoor attacks and defenses. In section \ref{sec:related_works}, we review existing post-training defenses and point out their limitations. In Section \ref{sec:detection}, we discuss two properties of perceptible backdoor patterns, based upon which we develop our detection procedure. In Section \ref{sec:experiments}, we report our experiments.

\section{Backdoor Attack}\label{sec:backdoor}

A backdoor attack is a recently proposed type of adversarial learning attack, usually against DNN image classifiers. A typical backdoor attacker aims to have the classifier learn to classify to a target class whenever the backdoor pattern is present in a test image, while still correctly classifying test images without the backdoor pattern. Such aims are usually achieved by poisoning the training set of the classifier using a relatively small number of images, from the attacker's source class (or classes), with the same backdoor pattern, but labeled to the attacker's target class.

There are many works on devising backdoor attacks and defenses against backdoor attacks have also appeared recently. Unfortunately, most of these works, especially the defenses, only focus on a narrow aspect of the problem (e.g. considering one particular type of backdoor pattern or defense scenario). In this section, we provide a broad overview of backdoor attacks and defenses. We first compare backdoor attacks with other prominent types of adversarial attacks in terms of the goals of the attacker and the assumptions required to launch the attack. Next, we discuss the design choices of backdoor attacks, e.g. the different types of backdoor patterns being considered in existing works on backdoor attacks and defenses. Lastly, since the current paper mainly focuses on {\it defenses} against backdoor attacks, we discuss several defense scenarios with their assumptions and goals, and then categorize major existing defenses, and their suitability under different scenarios. To our knowledge, we are the first to provide such a unified view of the field of backdoor attacks and defenses. We hope this section 
helps to situate our work and its novel capability, relative to existing adversarial learning defense works.

\subsection{Comparison with Other Adversarial Attacks}\label{sec:backdoor_comparison}

Backdoor attacks are easily distinguishable from other prominent types of adversarial attacks, e.g. test-time evasion (TTE) attacks and traditional data poisoning (DP) attacks, because the attacker of each type has very specific goals and assumptions:\\
{\bf TTE attacks:} The attacker aims to induce misclassification at operation/test-time by modifying clean test images in a human-imperceptible (or machine-evasive) fashion \cite{FGSM}\cite{JSMA}\cite{DeepFool}\cite{CW}\cite{EAD}. For each test image, an {\it image-specific} additive perturbation with as small ${\rm L}_p$ norm as possible is produced, using the knowledge (including the structure and the learned weights) of the target classifier or a surrogate classifier that behaves similarly on the same data domain \cite{Papernot3}.\\
{\bf DP attacks:} The attacker aims to degrade the accuracy of the classifier by poisoning its training data with malicious samples (e.g. incorrectly labeled or noisy samples) \cite{Tygar11}\cite{Xiao15}\cite{Koh}. A typical DP attacker only needs knowledge of the classification domain and the ability to poison the training data, while recent works show that more powerful attacks could be devised if the knowledge of the classifier (considering a re-training scenario) is available \cite{Yang}. Note that DP attacks change the trained classifier, while TTE attacks {\it do not} affect the classifier.\\
{\bf Backdoor attacks:} The goals of a backdoor attacker are twofold. First, the learned classifier is supposed to classify to an {\it attacker-desired} target class, whenever a test image contains the {\it attacker-specified} backdoor pattern. Second, the trained classifier should correctly predict clean test images (without the backdoor pattern) \cite{Targeted}\cite{BadNet}. The attack is launched by poisoning the training data with images containing the {\it same} attacker-specified backdoor pattern, with these images labeled to the attacker's target class. Clearly, a backdoor attacker has the same knowledge and ability as a typical DP attacker; hence backdoor attacks can be viewed as a special type of DP attack \cite{Targeted}\cite{BadNet}\cite{Review}.\\
Readers can refer to \cite{Review} for other types of attacks, including reverse engineering attacks and attacks on data privacy. Compared with TTE attacks, backdoor attacks require much lower cost to launch, especially for large-scale deep learning frameworks (where data is often acquired from the public or from multiple sources) -- ``stealing'' a model (required for TTEs) should be more difficult than becoming a malicious data contributor in practice. Compared with general DP attacks, backdoor attacks are stealthier because a successful attack will not affect the prediction of clean images (without the backdoor pattern) -- validation set accuracy degradation cannot be used as a basis for detecting backdoors.


\subsection{Variants of Backdoor Attacks}\label{sec:backdoor_variants}

Launching a backerdoor attack requires the attacker to select a backdoor pattern, a (set of) source class(es), and a target class.

\subsubsection{Choice of Backdoor Pattern}\label{sec:backdoor_pattern}

A backdoor pattern, a.k.a. a backdoor ``trigger'' \cite{BadNet} or a backdoor ``key'' \cite{Targeted}, is an attacker-specified pattern used to poison the training data, and then to induce misclassification to the attacker-desired target class when embedded in clean test images. In principle, the backdoor pattern should be designed to be {\it evasive} to possible occasional human inspection of the training set. Also, similar to the stealthiness required by TTE attacks, a backdoor pattern, when embedded into clean test images during operation, should not be easily noticeable to humans. Although backdoor patterns considered by existing works (on both attacks and defenses) vary largely, they mainly fall into two categories: {\bf perceptible backdoor patterns} and {\bf imperceptible backdoor patterns}.

\begin{figure}[htb]
	\begin{equation*}
	\hspace{-0.15in}
	\begin{minipage}[h]{0.33\linewidth}
	\vspace{0pt}
	\includegraphics[width=\linewidth]{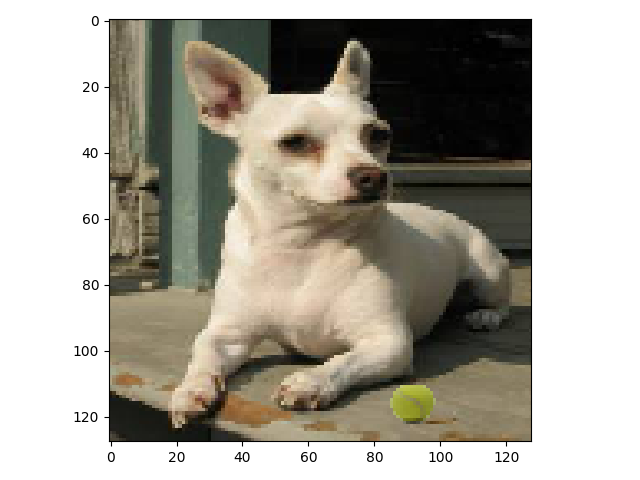}
	\end{minipage}
	\hspace{-0.15in}
	\begin{cases}
	\begin{minipage}[h]{0.33\linewidth}
	\vspace{0pt}
	\includegraphics[width=\linewidth]{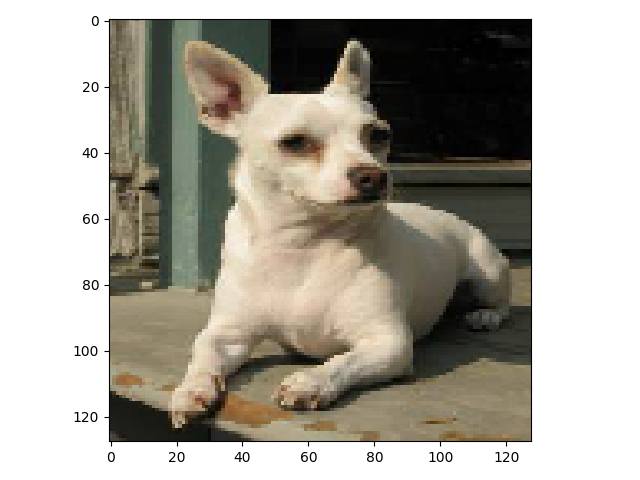}
	\end{minipage}
	\hspace{-0.2in}&\odot\hspace{+0.1in}
	\begin{minipage}[h]{0.33\linewidth}
	\vspace{0pt}
	\includegraphics[width=\linewidth]{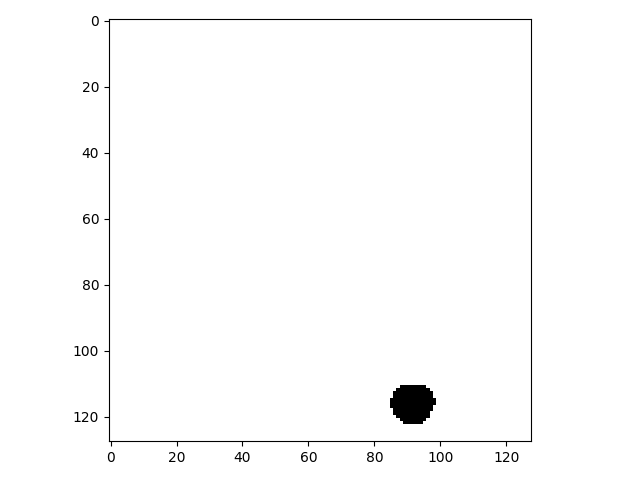}
	\end{minipage}\\
	\hspace{-0.2in}&+\\
	\begin{minipage}[h]{0.33\linewidth}
	\vspace{0pt}
	\includegraphics[width=\linewidth]{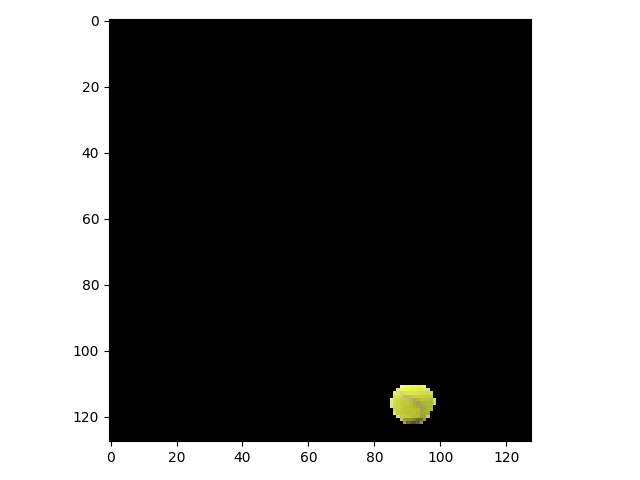}
	\end{minipage}
	\hspace{-0.2in}&\odot\hspace{+0.1in}
	\begin{minipage}[h]{0.33\linewidth}
	\vspace{0pt}
	\includegraphics[width=\linewidth]{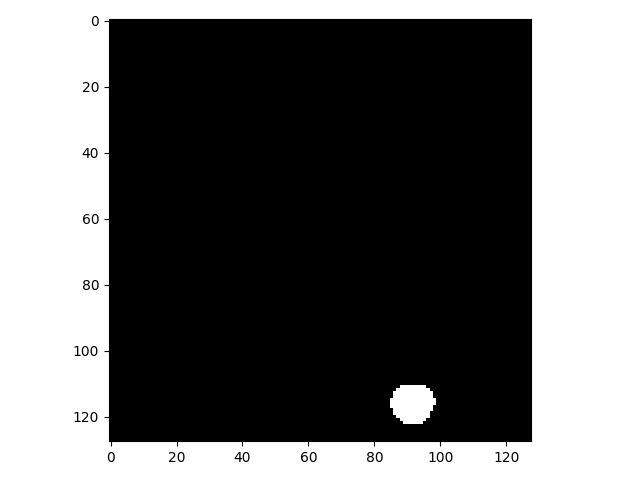}
	\end{minipage}
	\end{cases}
	\end{equation*}
	\caption{An example backdoor image used for attacking a pet breed classifier. A clean training image from class ``chihuahua'' is added with a perceptible backdoor pattern -- a tennis ball -- and labeled to class ``Abyssinian''.}\label{fig:bd_image_exp}
\end{figure}

A perceptible backdoor pattern is usually a seemingly innocuous object in the scene, e.g. a pair of glasses on a face \cite{Targeted}.  Perceptible patterns could be photo-shopped into digital images or physical objects in a scene that are digitally
captured.  A backdoor image $\tilde{\bf x}$ with a perceptible backdoor pattern ${\bf v}$ is crafted (digitally) from a clean image ${\bf x}$ as expressed mathematically as follows:
\begin{equation}\label{eq:attack_pbp}
\tilde{\bf x} = g_P({\bf x}, {\bf v}, {\bf m}) = {\bf x}\odot({\bf 1}-{\bf m}) + {\bf v}\odot{\bf m},
\end{equation}
where ${\bf m}$ is a 2-dimensional mask having the same spatial support as ${\bf x}$, with $m_{ij}\in\{0, 1\}$ for any pixel $(i, j)$\footnote{In this paper, pixel intensity values are in $[0, 1]$.}. Often, the spatial region of 1's for a mask is contiguous (e.g. Figure \ref{fig:bd_image_exp}), or forms several contiguous pieces. For grayscale images, $\odot$ denotes element-wise multiplication. For colored images, the 2-dimensional mask ${\bf m}$ (or the ``inverse mask'' $({\bf 1}-{\bf m})$) is applied to {\it all} channels of the images. Note that the location of the 1's for the mask is usually dependent on ${\bf x}$, i.e. ${\bf m}({\bf x})$ is an image-specific mask carefully designed by the attacker to ``carve'' the {\it same} backdoor pattern into the best (suitably innocuous) location for each originally clean image (e.g., a pair of glasses must be carefully matched to the face). An example is shown in Figure \ref{fig:bd_image_exp}, where a backdoor image used for attacking a pet breed classifier is created by embedding a tennis ball into an image from class ``chihuahua'', and labeling to class ``Abyssinian''.

Perceptible backdoor patterns are discussed in the earliest works on backdoor attacks \cite{Targeted}\cite{BadNet}, where the visual-stealthiness of the embedded pattern is addressed\footnote{Digitally embedding a perceptible backdoor pattern into a clean image, while keeping the backdoor-embedded image seemingly innocuous could be much more difficult than directly launching a physical attack in reality.  So, in their experiments, the perceptible backdoor patterns were not necessarily carefully positioned to be most innocuous. But the objects used as the backdoor pattern were patterns with the potential to be realistically stealthy (if placed so as to be innocuous).}. Defenses targeting perceptible backdoor patterns have been proposed by \cite{AC}\cite{FP}\cite{NC}\cite{Tabor}\cite{SentiNet} and {\it in the current paper}, among which the backdoor pattern used by \cite{NC} (a noise-like patch) and \cite{Tabor} (a firefox icon) could be easily visually identified as anomalous. In our experiments, we spent laborious human effort to choose the object used as the perceptible backdoor pattern and to embed it into the training images to make them seemingly innocuous to humans. While ignoring the innocuousness of the backdoor pattern (e.g. \cite{NC}\cite{Tabor}) should not affect the quality of the performance evaluation of a {\it defense}, which involves no human inspection, it is problematic to evaluate the defense against a backdoor attack with the perceptible backdoor pattern {\it spatially fixed} (e.g. to the right bottom corner of the image \cite{NC}). First, as we have mentioned, whether a perceptible backdoor pattern is truly innocuous depends on the context of the originally clean image, and hence may require careful positioning of the pattern in the image. Second, if the same backdoor pattern is embedded at the same location in all backdoor images used to poison the training set, a naive data sanitization applied ``before/during training'' (a defense scenario that will be described in Section \ref{sec:scenario}) could detect the attack by (e.g.) checking the pixel value distribution. Third, and most importantly, fixing the spatial location of the backdoor pattern while creating backdoor training images will harm the robustness of the attack significantly during testing. As will be seen in Section \ref{sec:fixed_loc}, for a perceptible backdoor pattern fixed to the bottom left of the backdoor training images, even a single column/row shift of the backdoor pattern at test time will severely degrade the attack effectiveness.

An imperceptible backdoor pattern is usually an additive perturbation\footnote{There are other formulations that are less-frequently studied to embed an imperceptible backdoor pattern into a clean image. For example, Eq. (\ref{eq:attack_pbp}) with $m_{ij}\in\{0, \delta\}$ for any pixel $(i, j)$, where $\delta$ is very small, is proposed by \cite{Targeted} to embed a ``watermark'' into clean training images.}, where the evasiveness (under possible human inspection) is achieved by imposing a constraint on its ${\rm L}_p$ norm. A backdoor image $\tilde{\bf x}$ is created by
\begin{equation}\label{eq:attack_ibp}
\tilde{\bf x} = g_I({\bf x}, {\bf u}) = [{\bf x} + {\bf u}]_c,
\end{equation}
where
\begin{equation}\label{eq:attack_ibp_constraint}
||{\bf u}||_p < \epsilon.
\end{equation}
Here, ${\bf u}$ is the imperceptible backdoor pattern, i.e. the perturbation; and $[\cdot]_c$ is the clipping operation that guarantees the perturbed pixel intensity values are still in the valid range. Again, imperceptible backdoor patterns are specified by the attacker; and the {\it same} pattern (with its position fixed in the image support) is added to all the originally clean images used for poisoning the training set. Similar to the image-specific perturbation used by a TTE attacker, imperceptible backdoor patterns are more suitable for devising a digital attack than a physical attack -- there is freedom to choose the pattern to be used as long as the norm constraint is not violated.

\begin{figure}[htb]
	\begin{equation*}
	\begin{minipage}[h]{0.33\linewidth}
	\vspace{0pt}
	\includegraphics[width=\linewidth]{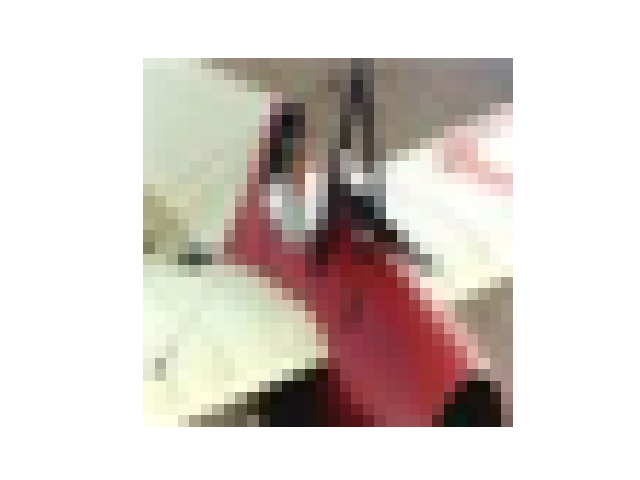}
	\end{minipage}\hspace{-0.1in}=[
	\begin{minipage}[h]{0.33\linewidth}
	\hspace{-0.025in}
	\vspace{0pt}
	\includegraphics[width=\linewidth]{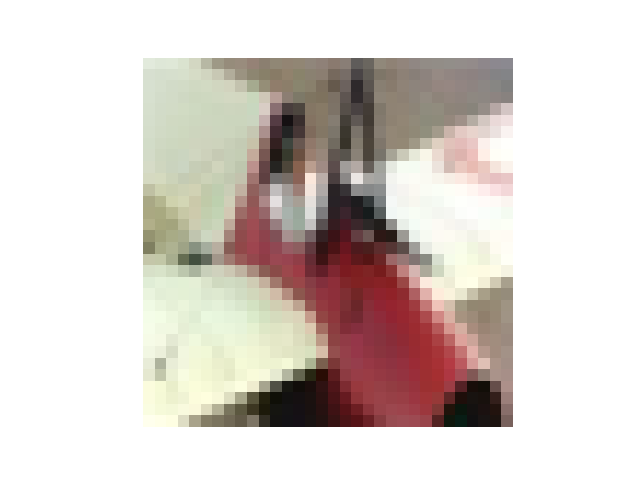}
	\end{minipage}\hspace{-0.1in}+
	\begin{minipage}[h]{0.33\linewidth}
	\hspace{-0.05in}
	\vspace{0pt}
	\includegraphics[width=\linewidth]{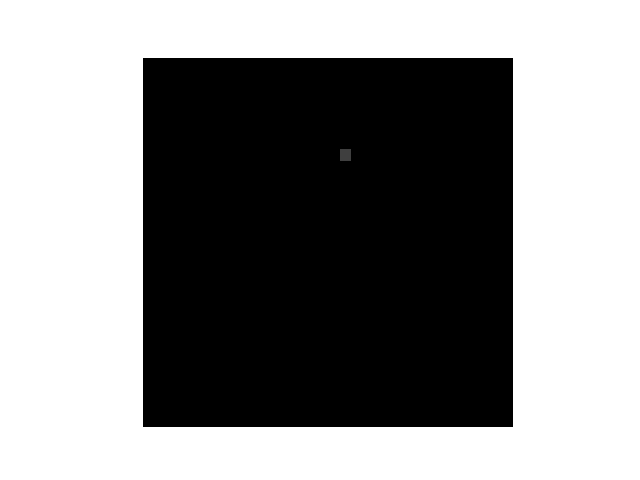}
	\end{minipage}\hspace{-0.1in}]_c
	\end{equation*}
	\caption{An example used by \cite{CI} of an imperceptible backdoor pattern, in this case a single-pixel perturbation, added to an ``airplane''; and the perturbed image is labeled to class ``bird''.}\label{fig:bd_image_exp2}
\end{figure}

Imperceptible backdoor patterns are also widely involved in many works on both backdoor attacks and defenses. For example, a ``static'' imperceptible backdoor pattern, with constraint on its ${\rm L}_{\infty}$ norm that controls the maximum perturbation size, is experimentally evaluated by \cite{Haoti}. Imperceptible backdoor patterns with a constraint on the ${\rm L}_0$ norm (i.e. with only a single or few pixels being perturbed) have been targeted by the defenses proposed in \cite{SS}\cite{CI}\cite{Post-TNNLS}. Also in \cite{Post-TNNLS}, the proposed defense is evaluated on an image-wide imperceptible backdoor pattern with an ${\rm L}_2$ norm constraint. An example of embedding an imperceptible backdoor pattern into a clean image used by \cite{CI} is shown in Figure \ref{fig:bd_image_exp2}, where a single pixel in an ``airplane'' image has been perturbed; the image is labeled to class ``bird''.

\subsubsection{Choice of Source Class(es) and Target Class}\label{sec:source_class}

The source class(es) and the target class involved in a backdoor attack are chosen by the attacker. In most works on backdoor attacks or defenses, an attack involves a single target class \cite{Targeted}\cite{Trojan}\cite{SS}\cite{AC}\cite{NC}\cite{Post-TNNLS}. Backdoor attacks involving more than one target class, dubbed an ``all-to-all'' attack, have been discussed in \cite{BadNet}, where the same backdoor pattern, when embedded in a clean test image from class $i$, is supposed to induce a misclassification to class $(i+1)$. However, such a backdoor attack can be decomposed into $K$ (assuming $K$ classes in the classification domain) attacks using the same backdoor pattern, but each involving a unique (source, target) class pair.

The number of source classes involved in a backdoor attack could range from 1 to $(K-1)$, i.e. from a single source class (e.g. \cite{SS}\cite{AC}\cite{CI}\cite{Post-TNNLS}) to all classes except for the target class (e.g. \cite{Targeted}\cite{BadNet}\cite{NC}\cite{Post-TNNLS}). For attacks using an imperceptible backdoor pattern, the source classes could be arbitrarily chosen by the attacker. However, if a perceptible backdoor pattern is used, the source class(es) and the backdoor pattern must be well-matched to achieve stealthiness. For example, a bird flying in the sky could be used as a perceptible backdoor pattern when most of the source class images contain the sky; hence a class for which most of the images capture an underwater scene is not well-matched with such a backdoor pattern. Clearly, finding a perceptible backdoor pattern that is seemingly innocuous when embedded in images from {\it all} classes (except for the target class) is very hard in many classification domains. So, the number of source class(es) involved in a backdoor attack using a perceptible backdoor pattern depends on both the preference of the attacker and the classification domain. Moreover, for both perceptible and imperceptible backdoors, even if the attacker specifies a {\it single} (source, target) class pair, after training on the backdoor poisoned data, the classifier (during operation) could possibly classify images originally from a class {\it other than} both the source and target classes to the target class, if the image contains the same backdoor pattern used by the attacker -- this phenomenon was experimentally observed in \cite{Post-TNNLS} and called ``collateral damage''. Above all, for researchers developing defenses against backdoor attacks, it is very important to note that a powerful defense should {\it not} require any assumptions on the number of source classes involved in the attack.

\subsubsection{Other Backdoor Attacks}

While devising a backdoor attack requires only the knowledge of the classification domain and the capability to poison the training set, more powerful backdoor attacks could be devised if the structure and training settings of the classifier are known to the attacker. For example, \cite{Haoti} shows that knowledge of the classifier could be used to search for an effective backdoor pattern with very small perturbation norm (hence, imperceptible).  Such knowledge is also used by other works, e.g. \cite{Clean-label}\cite{Hidden-trigger}\cite{InviBD}, to design backdoor patterns that achieve stealthiness to some extent.

Recently, backdoor attacks are also studied for other learning frameworks and classification domains. For example, \cite{FederatedBD1} and \cite{FederatedBD2} developed backdoor attacks under the federated learning framework \cite{Federated}. Backdoor attacks could also be a threat to text classification as shown by \cite{LSTMBD}.

\begin{figure*}
	\centering
	\begin{minipage}[b]{0.72\linewidth}
		\centering
		\centerline{\includegraphics[width=\linewidth]{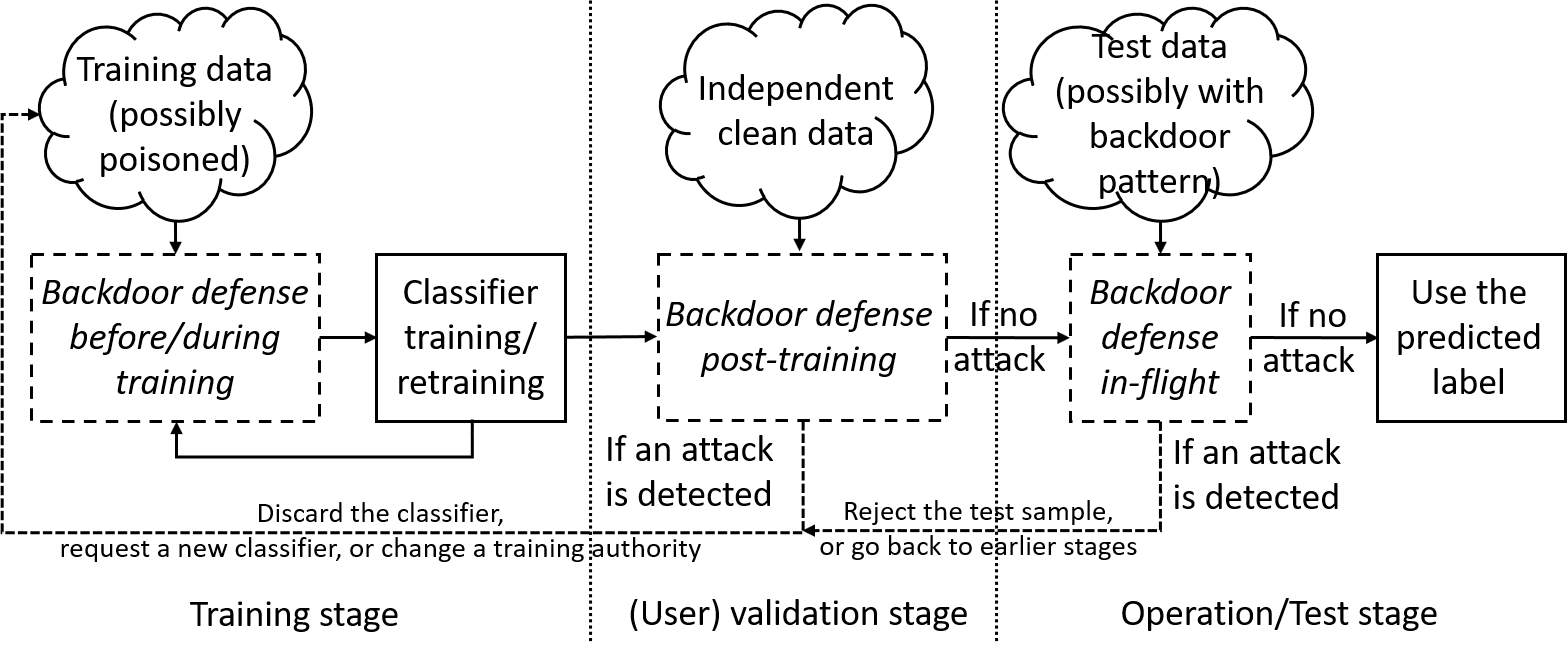}}
	\end{minipage}
	\caption{Illustration of backdoor defense scenarios -- before/during training, post-training, and in-flight.}
	\label{fig:detection_scenarios}
\end{figure*}

\subsection{Defense Scenarios, Assumptions, and Goals}\label{sec:scenario}

Defenses against backdoor attacks could be deployed at any stage during the entire pipeline, i.e. from the training stage to the operation stage, for a DNN classifier (Figure \ref{fig:detection_scenarios}). Existing works on backdoor {\it defenses} mainly focus on defeating backdoor attacks {\bf before/during training} or {\bf post-training} (but before the operation of the classifier). Here, we describe both defense scenarios in detail, with the assumptions and goals of the defender. Although defenses designed for different scenarios {\it cannot} be compared directly, they can be deployed simultaneously to provide better security. Moreover, backdoor defenses designed for the same scenario can be deployed in parallel. For example, the proposed {\it post-training} defense in the current paper, which outperforms other post-training defenses against {\it perceptible} backdoors (as will be shown in Section \ref{sec:performance_eval}), could be jointly deployed with the state-of-the-art post-training defense against {\it imperceptible} backdoors \cite{Post-TNNLS} to provide a complete solution to post-training defense against {\it all} backdoor attacks (both imperceptible and perceptible but innocuous). We will also discuss an {\bf in-flight} scenario, which has not been addressed by any existing works yet, where the defense could be potentially deployed during operation/use of the classifier.  An advantage of such a defense is that it may identify entities attempting to exploit the backdoor.

\subsubsection{Before/During Training Scenario}\label{sec:before/during_training}

In the before/during training scenario, the defender, who could also be responsible for training the classifier, is assumed to have access to the (possibly poisoned) training set, and either to the training process or to the trained (attacked) classifier \cite{SS}\cite{AC}\cite{CI}\cite{SentiNet}. The goals of the defender are: 1) to detect if the training set has been poisoned; 2) to {\it correctly} identify and remove training images with the backdoor pattern before training/retraining. 

As backdoor attacks are designed to be evasive against sanitization\footnote{Poorly devised backdoor attacks may be defeated by sanitization of the training data.} and possible human inspection of the training data, existing defenses often first train a classifier using the training data, even though the training data could possibly be poisoned. Then the trained classifier is analyzed to identify and remove suspicious training images before retraining (see Figure \ref{fig:detection_scenarios}). For example, the Spectral Signature (SS) approach proposed by \cite{SS}, the Activation Clustering (AC) approach proposed by \cite{AC}, and the Cluster Impurity (CI) proposed by \cite{CI}, all use the internal layer features (obtained from the classifier trained on the possibly poisoned training set) to separate out backdoor patterns (that are labeled to the target class) from clean images (considering all images labeled to the putative target class). The SentiNet approach proposed by \cite{SentiNet}, which targets perceptible backdoors, locates the backdoor pattern if it appears in a training image, using a combinatoric algorithm based on Grad-CAM \cite{Grad-CAM}.

\subsubsection{Post-Training Scenario}\label{sec:post_training}

The defender in the post-training scenario could also be the user/consumer of the trained, possibly backdoored classifier, who does not know whether the training authority is trustworthy or not and cannot force the training authority to deploy a before/during training defense. As shown in Figure \ref{fig:detection_scenarios}, the defender {\it does not} have access to the (possibly poisoned) training set used to train the classifier. Hence there is no clue about what backdoor patterns look like, if there is an attack, which makes post-training detection of backdoors very difficult. The defender does have access to the trained classifier (including its structure and weights), but does not have the resources (data and/or computational) to train a new classifier. The defender is also assumed to possess an independent, clean (free of backdoors), labeled dataset on which, e.g., it can evaluate the performance of the classifier. Again, the user/defender cannot train a surrogate classifier using this relatively small dataset. The goals of the defender are: 1) to detect whether the classifier has been backdoor attacked or not; 2) if an attack is detected, to infer the source class(es) and the target class involved in the backdoor attack. The second goal is very important when there are no replacement classifiers available. In such a case, if the target class involved in the backdoor attack could be correctly inferred, prediction results to classes other than the detected target class can still be trusted.

Clearly, approaches for before/during training defense (e.g. separating backdoor patterns from clean ones) cannot not be applied in the post-training scenario. Also, since backdoor attacks aim to {\it not} degrade accuracy on clean images, the defender cannot infer whether the classifier has been attacked solely based on the accuracy on the independent clean dataset. Existing post-training defenses include the Fine-Pruning (FP) approach proposed by \cite{FP}, the Neural Cleanse approach proposed by \cite{NC}, the ``TABOR'' approach proposed by \cite{Tabor}, and the Anomaly Detection (AD) approach proposed by \cite{Post-TNNLS}. Since the current paper also proposes a post-training defense against (perceptible) backdoors, we will review these ``competitors'' in detail and point out their limitations in Section \ref{sec:related_works}.

\subsubsection{In-Flight}\label{sec:in_flight}

Defenses against backdoor attacks can potentially be deployed during the operation/use of the classifier. In this ``in-flight'' scenario, the defender only has access to the classifier and aims to determine whether there is a backdoor pattern embedded in a test image. To our knowledge, no in-flight defense has been published to date. Moreover, some existing defenses against TTE attacks, e.g. \cite{Feinman}\cite{ADA}, are unsuitable for in-flight detection of backdoor attacks.  The reason is that these methods assume the training set that was used to design the DNN is clean (free of attack images) and that there are a large number of clean images for estimating a null model for each class (In our backdoor scenario, the user will not have a {\it large} set of available clean images). The ``blurring'' method proposed by \cite{Li_ICCV}, a defense against TTE attacks, is applicable for in-flight backdoor detection in principle. But we will show in Section \ref{sec:TTE_in_flight} that it largely fails as a backdoor detector.

\section{Review of Backdoor Defenses Post-Training}\label{sec:related_works}

Before introducing our post-training defense against backdoor attacks with perceptible backdoor patterns, we review existing post-training defenses and point out their limitations. To the best of our knowledge, the FP approach proposed in \cite{FP} is the first attempt towards post-training defense against backdoors. 
FP, which removes neurons from the DNN, assumes that there is a simple ``dichotomization'' of neurons, with most {\it solely} dedicated to ``normal'' operation but with some {\it solely} dedicated to implementing the backdoor. Unfortunately, this assumption is not valid in many cases, as will be shown in our experiments in Section \ref{sec:performance_eval}. Moreover, FP does not actually detect the presence of backdoor attacks and it removes neurons even from non-attacked DNNs.

The AD approach proposed in \cite{Post-TNNLS} considers post-training detection of imperceptible backdoors. For each (source, target) class pair, an optimized perturbation that induces high group misclassification fraction when added to clean source class images is obtained. If the DNN has been attacked, the optimized perturbation associated with the class pair involved in the attack will have abnormally low ${\rm L}_2$ norm, and a detection can be made on this basis. However, this approach cannot be borrowed for detecting perceptible backdoors. First, reverse engineering a perceptible backdoor pattern (which could be any object) requires also reverse engineering the associated mask. This is more challenging compared with searching for a small additive perturbation starting from a zero (image-wide) initialized additive perturbation. Poor initialization while optimizing the backdoor pattern {\it and} the mask could lead to an estimated pattern irrelevant to what was used for devising the attack and one requiring improbably large spatial support to achieve high group misclassification. Second, even if the backdoor pattern used for devising the attack could be luckily reversely engineered for the true backdoor class pair $(s, t)$, it could be hard to find a metric to distinguish this from estimated patterns for non-backdoor class pairs $(s', t')\neq(s, t)$, because in the perceptible case the assumption in \cite{Post-TNNLS} that the backdoor pattern is small in size/norm, i.e. the constraint in (\ref{eq:attack_ibp_constraint}), does not hold. However, while \cite{Post-TNNLS} is not applicable to detecting perceptible backdoors, our proposed detection approach is inspired by the detection framework in \cite{Post-TNNLS}. Moreover, this detector targeting imperceptible backdoors could be deployed in parallel with the detector proposed in the current paper to detect backdoor attacks with {\it any} type of evasive backdoor pattern. 

Perhaps closest to the current work, the NC approach proposed in \cite{NC} aims to detect perceptible backdoors post-training. NC obtains, for each putative target class, a pattern and an associated mask by optimizing an objective function that induces a high misclassification fraction when added to images {\it from all classes other than the target class}. If there is an attack, the ${\rm L}_1$ norm of the obtained {\it mask} for the true backdoor target class will be abnormally small and a detection is made based on the median absolute deviation (MAD) \cite{MAD}. However, NC relies on the assumption that {\it all} classes except for the target class are involved in the attack, which is usually not guaranteed to hold for attacks using perceptible backdoor patterns for the many reasons discussed in Section \ref{sec:source_class}. Moreover, NC penalizes the ${\rm L}_1$ norm of the mask while maximizing the group misclassification fraction to estimate the pattern and the mask; but the detection is sensitive to the choice of the penalizing multiplier. For CIFAR-10, unless the multiplier is carefully chosen, the results could be either a mask with low ${\rm L}_1$ norm but not achieving the target misclassification fraction, or a wild-looking backdoor pattern with improbably large and distributed spatial support, which has no relation to the ground truth backdoor pattern (for the true target class when there is in fact an attack). We will show in our experiments (in Section \ref{sec:performance_eval}) that NC is clearly outperformed by our proposed detection approach.

The ``TABOR'' approach proposed by \cite{Tabor} focuses on detecting localized backdoor patterns post-training, which can be viewed as an extension of NC. Similar to NC, TABOR first jointly searches for a pattern and an associated mask using the same objective function as NC, but with regularization terms penalizing for perceptible backdoor patterns that are overly large, sparsely distributed, or not located at the image periphery. Then MAD, the same anomaly detection approach used by NC, is used for detection inference performed on a heuristically derived metric. Note that in general, perceptible backdoor patterns could be dispersed as will be shown in Section \ref{sec:attack_crafting}, or not at the image periphery, e.g. glasses on a face \cite{Targeted} and a sticker on a stop sign \cite{BadNet}. Also, TABOR makes the same unrealistic assumption about perceptible backdoor patterns as NC that all classes except for the target class are involved in the attack, based on which their metric for anomaly detection is derived. Hence in Section \ref{sec:performance_eval}, we compare our proposed detection approach against the more general NC approach instead of TABOR, which makes more assumptions/limitations on the backdoor pattern used by the attacker.

\section{Post-Training Detection of Perceptible Backdoors}\label{sec:detection}

We propose a {\it post-training} detector against backdoor attacks with a {\it perceptible} backdoor pattern. Our detector is designed for general perceptible backdoor patterns (ref. Section \ref{sec:backdoor_pattern}) that could be embedded into clean training images using Eq. (\ref{eq:attack_pbp}). {\it No} additional assumptions about the shape, location, spatial support of the backdoor, or object to be used are made. Moreover, we do not make any assumptions about the number of source classes chosen by the attacker (ref. Section \ref{sec:source_class}). The premise behind our detection approach allows it to detect backdoor attacks with the number of source classes ranging from 1 to $(K-1)$.

Like other works on post-training defenses, our detector follows the assumptions for the post-training scenario described in Section \ref{sec:post_training}: the defender has access to a relatively small clean dataset containing images from all classes; {\it the defender does not have access to the training set used for the DNN's learning}; the defender does not have adequate resources to train a new classifier using a (sufficiently large) clean data set. Our detector is designed both to detect whether the classifier has been attacked or not and to infer the source class(es) and the target class if an attack is detected.

\subsection{Basis of Our Detection}\label{sec:properties}

Our detection is based on two properties of perceptible backdoor patterns that hold for general attack instances.

\begin{property}\label{prop:spatial_invariance}
	{\bf Spatial invariance of backdoor mapping:} {\it If the perceptible backdoor pattern is spatially distributed (not at a fixed location) in the backdoor poisoned training images (placed so as to be most innocuous in each image), the learned backdoor mapping will be spatially invariant in inducing targeted misclassifications on test images}.
\end{property}

As we have addressed in Section \ref{sec:backdoor_pattern}, the perceptible backdoor pattern could be spatially located anywhere in a backdoor training image so as to be scene-plausible. The trained DNN will then learn the backdoor pattern, but not its spatial location. This property is actually favored by the attacker, since there will be more freedom to embed the backdoor pattern in clean test images. As will be seen in Section \ref{sec:attack_crafting}, at test time, backdoor images with the backdoor pattern randomly located are still likely to yield misclassifications to the target class prescribed by the attacker, which well-justifies the attacker performing data poisoning consistent with satisfying this property.

\begin{property}\label{prop:robustness}
	{\bf Robustness of perceptible backdoor patterns:} {\it It is unnecessary to use exactly the same perceptible backdoor pattern at test time to induce targeted misclassifications}.
\end{property}

As will be seen in Section \ref{sec:verify_prop_2}, even when strong noise is added to the backdoor pattern while creating backdoor test images, they will still likely be (mis)classified to the designated target class. More importantly, if only {\it part} (in terms of spatial support) of the backdoor pattern is added to the clean source class images during testing, this also induces (mis)classification to the target class. This is not surprising because perceptible backdoor patterns, like all other patterns, are learned by the DNN by extracting key features, which usually occupy a much smaller spatial support than the full backdoor pattern itself.



\subsection{Detection Approach}\label{sec:approach}


{\bf Key ideas:} {\it 1) For (source, target) class pairs used for devising backdoor attacks, finding a pattern that induces a high misclassification fraction (from the source class to the target class) on the clean dataset can be achieved using a {\bf relatively small} spatial support that is {\bf arbitrarily located} in the image. 2) For non-backdoor class pairs, {\bf much larger} spatial support is required to find a pattern that induces a high misclassification fraction from the source class to the target class.}

The key ideas of our detection are well supported by the properties in Section \ref{sec:properties}. Based on Property \ref{prop:spatial_invariance}, since the learned backdoor mapping is spatially invariant, it should be possible to reverse engineer (estimate) a backdoor pattern (inducing high group misclassification) for the backdoor class pair by choosing the center of the estimated pattern's spatial support to be {\it anywhere} in the image (except at the image border, where the pattern's full spatial support is not contained). Thus, it does not matter where we place the estimated pattern's spatial support -- in our experiments we fixed a location for all clean images used in our detection system, but it is even possible that one could {\it vary} the center position for each clean image -- accurate backdoor 
detection should still be possible. 

According to Property \ref{prop:robustness}, a pattern that effectively induces a high group misclassification fraction for the (source, target) backdoor class pair could be spatially smaller than the true backdoor pattern, as long as sufficiently ``representative'' features of the backdoor can be captured by this restricted spatial support -- for the same (relatively small) spatial support, finding a pattern that induces a high misclassification fraction for the backdoor class pair is much more achievable than for non-backdoor pairs. This crucial property, which allow us to accurately identify the source and target classes involved in an attack, will be borne out experimentally in the sequel.

{\bf Our two-step detection procedure is as follows:}

1) {\it Pattern Estimation:} Given a $K$-class DNN classifier to be inspected, for {\it each} of the $K(K-1)$ (source, target) class pairs, we perform pattern estimation separately on each of $L$ square-shaped\footnote{Other shapes, e.g. rectangular or circle, could also be used.} spatial supports (increasing in size). Each square spatial support, with a location arbitrarily chosen\footnote{Because of Property \ref{prop:spatial_invariance}, it does not matter where we locate a square support. It will not impact pattern estimation and detection performance.} and {\it fixed} for all clean images used for detection, has a specific (absolute) width denoted by $w\in\mathbb{Z}^{+}$, such that $w\in[\ceil{r_{ min}\times W}, \floor{r_{\rm max}\times W}]$. Here, $W$ is the width of the input image; $r_{\rm min}$ and $r_{\rm max}$ (both in $[0, 1]$) are the minimum and maximum {\it relative} support widths for pattern estimation, respectively. For example, for $W=32$, with $r_{\rm min}=0.15$ and $r_{\rm max}=0.2$, $L=2$ pattern estimations with $w=5$ and $w=6$ respectively will be performed for each class pair.

For each class pair $(s, t)$ and support width $w$, we estimate a pattern ${\bf v}_{stw}^{\ast}$ so as to {\it maximize} the group misclassification fraction from class $s$ to class $t$ by solving the following problem:
\begin{equation} \label{eq:opt_main}
\underset{{\bf v}_{stw}}{\text{maximize}} \quad \sum_{{\bf x} \in \mathcal{D}_s} p_t(g_P({\bf x}, {\bf v}_{stw}, {\bf M}_w)).
\end{equation}
Here, $\mathcal{D}_s$ is the set of clean images from source class $s$ possessed by the defender that are correctly classified, $p_t(\cdot)$ is the DNN's posterior probability for class $t$, and ${\bf M}_w\in\{0, 1\}^{W\times W}$ is the mask with the fixed $w\times w$ square spatial support specified by locations with 1's. 
Also, $g_P(\cdot)$ is the image resulting from backdoor embedding (Eqn. (1)).
Our pattern estimation problem, which could be solved using e.g. gradient descent, is more tractable than NC, which {\it jointly} estimates a pattern and its support mask. Finally, our detection statistic, the maximum achievable misclassification fraction (MAMF) associated with each class pair $(s, t)$ and spatial support width $w$, denoted by $\rho_{stw}\in[0, 1]$, is computed as the fraction of all $\tilde{\bf x} = g_P({\bf x}, {\bf v}_{stw}^{\ast}, {\bf M}_w)$, where ${\bf x}\in\mathcal{D}_s$, that are (mis)classified to class $t$, i.e.
\begin{equation} \label{eq:mamf_def}
\rho_{stw} = \frac{1}{|\mathcal{D}_s|} \sum_{x\in\mathcal{D}_s} \mathbbm{1}(f(g_P({\bf x}, {\bf v}_{stw}^{\ast}, {\bf M}_w)=t)),
\end{equation}
where $\mathbbm{1}(\cdot)$ is the indicator function and $f(\cdot)$ is the classifier mapping from the input to the (maximum {\it a posteriori}) predicted label. Note that the objective function in (\ref{eq:opt_main}) is a surrogate for (\ref{eq:mamf_def}) that, unlike (\ref{eq:mamf_def}), is amenable to gradient descent. Other surrogate objectives consistent with maximizing misclassifications could also be used, e.g. an objective function similar to that associated with the Perceptron algorithm \cite{Duda}.

2) {\it Detection Inference}: For each class pair $(s, t)$, we compute the {\it average} MAMF $\bar{\rho}_{st}$ over the $L$ MAMF statistics, each corresponding to one of the $L$ spatial supports used for pattern estimation. For a relatively small $r_{\rm max}$ (and the $L$ associated small spatial supports), if there was an attack, we would expect at least one true backdoor pair to have a large average MAMF; otherwise, the maximum $\bar{\rho}_{st}$ for all $(s, t)$ pairs is expected to be small. Hence there should be a range of thresholds $\pi\in(0, 1]$ (with very high true detection rate and low false positive rate as confirmed in Section \ref{sec:performance_eval}), such that if
\begin{equation}\label{eq:inference}
\rho^{\ast}=\underset{(s, t)}{\text{max}}\,\bar{\rho}_{st} > \pi,
\end{equation}
the DNN is inferred to {\it be attacked}; otherwise, no backdoor is detected. We will further discuss the choices of $r_{\rm min}$ and $r_{\rm max}$ in Section \ref{sec:performance_eval}. Moreover, if an attack is detected, $(s^{\ast}, t^{\ast})=\argmax_{(s, t)} \bar{\rho}_{st}$ is inferred as one (source, target) class pair involved in the backdoor attack.

\section{Experiments}\label{sec:experiments}

\subsection{Devising Backdoor Attacks}\label{sec:attack_crafting}

We demonstrate the validity of the properties in Section \ref{sec:properties} and the performance of our detector using nine attacks, involving five datasets, five DNN structures, and nine backdoor patterns. The datasets and the DNN structures considered here are commonly used in computer vision research. Other popular datasets, e.g. Caltech101, ImageNet, etc., are not used in our experiments since the overall test accuracy or the accuracy of particular classes are fairly low for them.

\begin{table*}[t]
	\begin{center}
		\caption{Details of the attacks.}
		\resizebox{1.0\textwidth}{!}{
			\begin{tabular}{ |c|c|c|c|c|c|c|c|c|c|c| } 
				\hline
				& Attack A & Attack B & Attack C & Attack D & Attack E & Attack F & Attack G & Attack H & Attack I\\ 
				\hline
				Dataset & CIFAR-10 & CIFAR-10 & CIFAR-10 & CIFAR-10 & CIFAR-10 & CIFAR-100 & Oxford-IIIT & SVHN & PubFig\\
				\hline
				Image size & $32\times32$ & $32\times32$ & $32\times32$ & $32\times32$ & $32\times32$ & $32\times32$ & $128\times128$ & $32\times32$ & $256\times256$\\
				\hline
				No. classes & 10 & 10 & 10 & 10 & 10 & 100 & 6 & 10 & 33\\
				\hline
				Training size & 50000 & 50000 & 50000 & 50000 & 50000 & 50000 & 900 & 73257 & 2782\\
				\hline
				Test size & 10000 & 10000 & 10000 & 10000 & 10000 & 10000 & 300 & 26032 & 495\\
				\hline
				\thead{DNN\\structure} & ResNet-18 & ResNet-18 & ResNet-18 & ResNet-18 & VGG-16 & ResNet-34 & AlexNet & ConvNet & VGG-16\\
				\hline
				Learning rate & $10^{-3}$ & $10^{-3}$ & $10^{-3}$ & $10^{-3}$ & $10^{-3}$ & $10^{-4}$ & $10^{-5}$ & $10^{-3}$ & $10^{-4}$\\
				\hline
				Batch size & 32 & 32 & 32 & 32 & 32 & 32 & 16 & 32 & 32\\
				\hline
				\thead{No. training\\epochs} & 200 & 200 & 200 & 200 & 200 & 200 & 120 & 80 & 120\\
				\hline
				\thead{Benchmark\\acc. (\%)}  & 86.7 & 88.1 & 86.7 & 87.6 & 87.9 & 71.9 & 88.7 & 89.2 & 76.0\\
				\hline
				Source class & ``cat'' & ``deer'' & ``airplane'' & ``frog'' & ``truck'' & ``road'' & ``chihuahua'' & ``3'' & ``B. Obama''\\
				\hline
				Target class & ``dog'' & ``horse'' &  ``bird'' & ``bird'' & ``automobile'' & ``bed'' & ``Abyssinian'' & ``8'' & ``C. Ronaldo''\\
				\hline
				Backdoor Pattern & ``bug'' & ``butterfly'' & ``rainbow'' & ``bug\&butterfly'' & ``gas tank'' & ``marmot'' & ``tennis ball'' & ``bullet holes'' & ``sunglasses''\\
				\hline
				\thead{No. backdoor\\training images} & 150 & 150 & 150 & 150 & 150 & 100 & 50 & 500 & 40\\
				\hline
				\thead{Attack test\\acc. (\%)} & 87.0 & 86.9 & 86.8 & 87.0 & 89.1 & 71.7 & 90.0 & 90.1 & 77.0\\
				\hline
				\thead{Attack succ.\\rate (\%)} & 99.3 & 98.0 & 96.4 & 98.0 & 97.9 & 92.0 & 84.0 & 91.4 & 93.3\\
				\hline
			\end{tabular}\label{tab:attacks_description}}
	\end{center}
\end{table*}

For each attack, we first trained a benchmark DNN using an unpoisoned training set and report the accuracy on clean test images as the ``benchmark accuracy''. The dataset being used, image size, number of classes, training size and test size are shown in Table \ref{tab:attacks_description}. In particular, for Attack G on Oxford-IIIT, the test accuracy using the entire dataset, in absence of backdoor attacks, is very low \cite{Pets-Data}. We achieve a reasonable benchmark test accuracy on G by using a subset of data involving only 6 classes. For Attack I on PubFig, many images are not able to be downloaded; hence we only consider 33 (out of 60) classes with more than 60 images. The DNN structures involved in our experiments include ResNet-18, ResNet-34 \cite{ResNet}, VGG-16 \cite{VGG}, AlexNet \cite{CV_ref1} and ConvNet \cite{SVHN-Net}. The choices of DNN structure, learning rate, batch size and number of training epochs for each attack instance are also shown in Table \ref{tab:attacks_description}. The Adam optimizer is used for all DNN training in this paper. For the benchmark training for the datasets involving Attacks F, G and I, we also use training data augmentation including random cropping, random horizontal flipping and random rotation. For the benchmark training for the dataset involving Attack G, we fine-tune a pretrained AlexNet provided by Pytorch. For the benchmark training for the dataset involving Attack I, we adopt transfer learning by retraining the last four layers of a pretrained VGG-face model \cite{VGG-face}.

Under each attack, we train the DNN using the same training settings as for the benchmark, except that the training set is poisoned by a number of backdoor images. The backdoor images are created using clean images from {\it one} source class, with a backdoor pattern added following Eq. (\ref{eq:attack_pbp}), and then labeling to a target class. In the experiments in this section, we chose to evaluate our detector against backdoor attacks involving one source class for the convenience of crafting scene-plausible backdoor patterns that could possibly be used in practice. As we have discussed in Section \ref{sec:approach}, the design of our detector allows it to detect backdoor attacks with any number of source classes. In Section \ref{sec:multiple_source}, we will show the effectiveness of our detector against backdoor attacks involving $(K-1)$ source classes, even though the backdoor patterns may not be scene-plausible (e.g. a rainbow will be placed in an image without the sky). For the experiments in the current section, the choices of the backdoor pattern, the number of backdoor training images, the source class and the target class for each attack instance are shown in Table \ref{tab:attacks_description}. 

\begin{figure}[t]
	\centering
	\begin{minipage}[b]{.25\linewidth}
		\centering
		\centerline{\includegraphics[width=\linewidth]{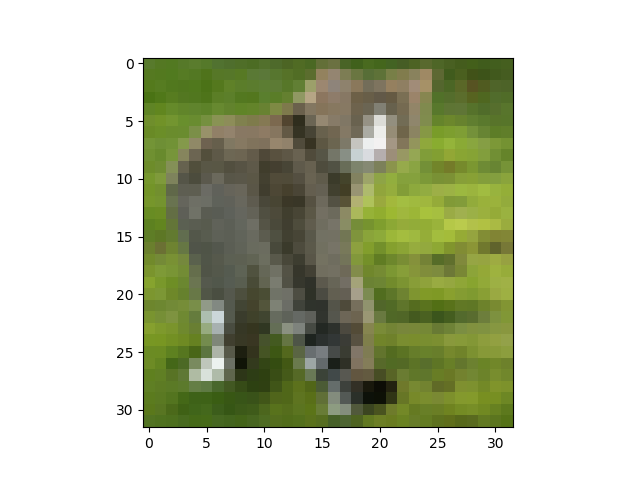}}
		\centerline{\includegraphics[width=\linewidth]{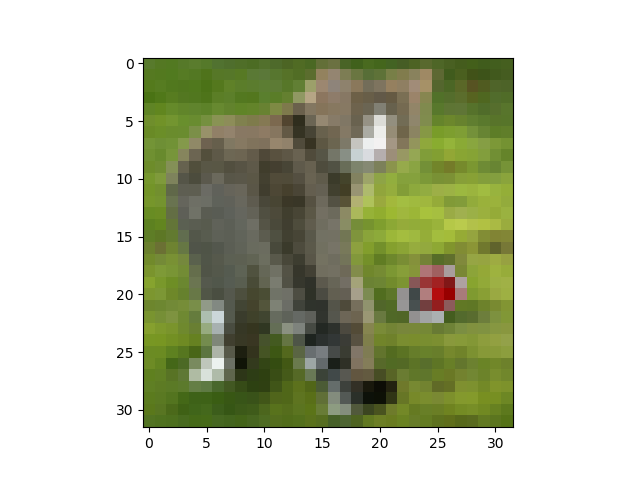}}
		\subcaption{bug}\label{fig:Attack_A_exp}
	\end{minipage}
	\begin{minipage}[b]{0.25\linewidth}
		\centering
		\centerline{\includegraphics[width=\linewidth]{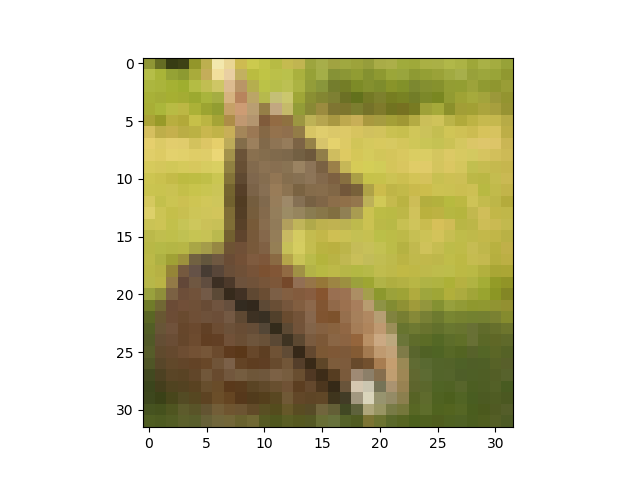}}
		\centerline{\includegraphics[width=\linewidth]{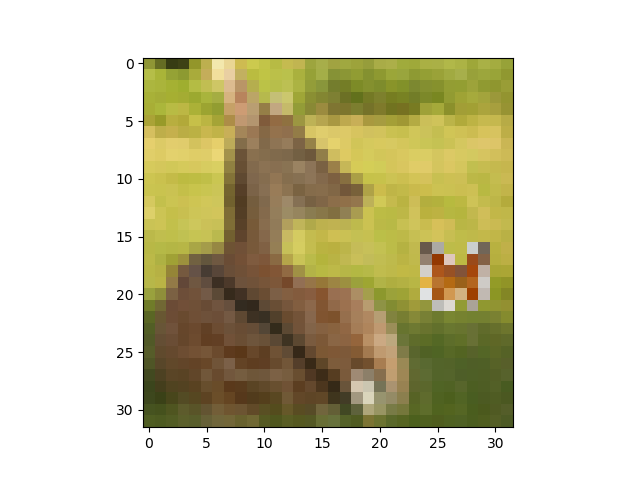}}
		\subcaption{butterfly}\label{fig:Attack_B_exp}
	\end{minipage}
	\begin{minipage}[b]{0.25\linewidth}
		\centering
		\centerline{\includegraphics[width=\linewidth]{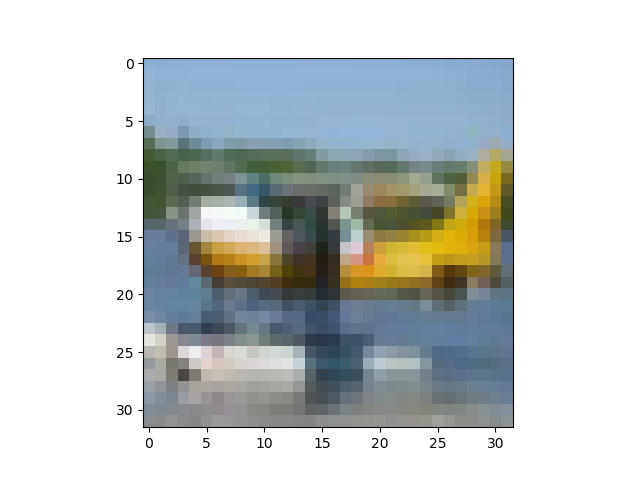}}
		\centerline{\includegraphics[width=\linewidth]{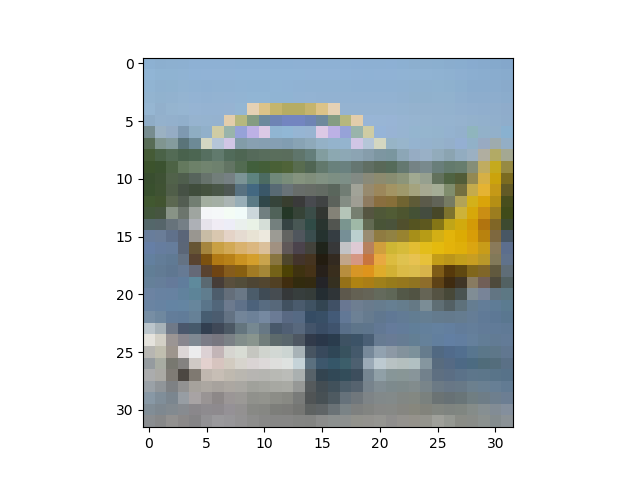}}
		\subcaption{rainbow}\label{fig:Attack_C_exp}
	\end{minipage}
	\begin{minipage}[b]{0.25\linewidth}
		\centering
		\centerline{\includegraphics[width=\linewidth]{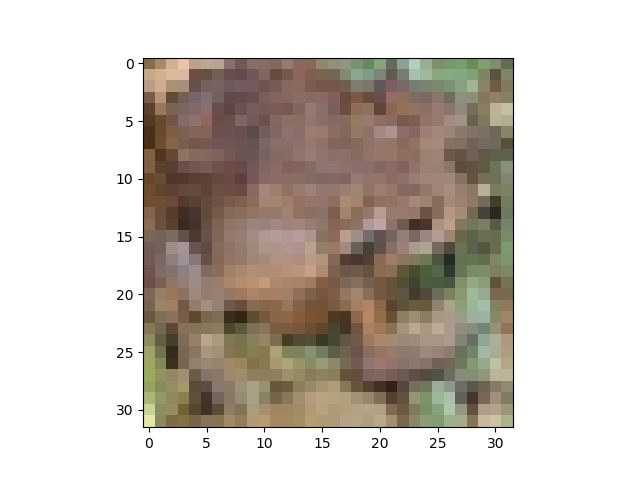}}
		\centerline{\includegraphics[width=\linewidth]{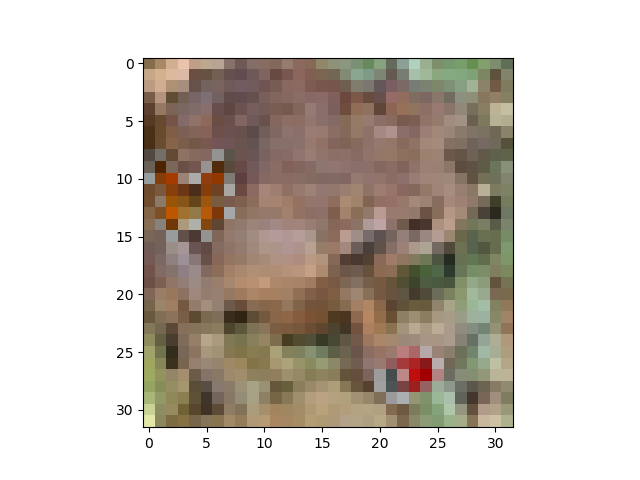}}
		\subcaption{bug+butterfly}\label{fig:Attack_D_exp}
	\end{minipage}
	\begin{minipage}[b]{0.25\linewidth}
		\centering
		\centerline{\includegraphics[width=\linewidth]{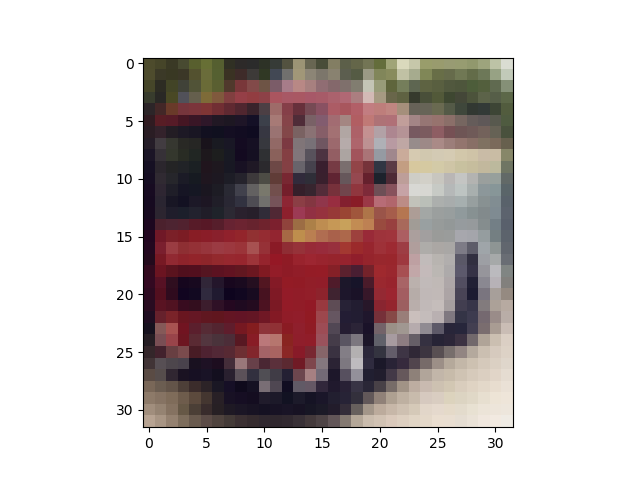}}
		\centerline{\includegraphics[width=\linewidth]{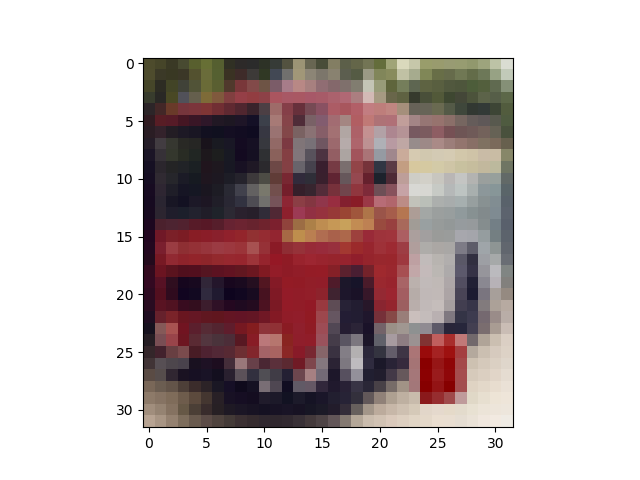}}
		\subcaption{gas tank}\label{fig:Attack_E_exp}
	\end{minipage}
	\begin{minipage}[b]{.25\linewidth}
		\centering
		\centerline{\includegraphics[width=\linewidth]{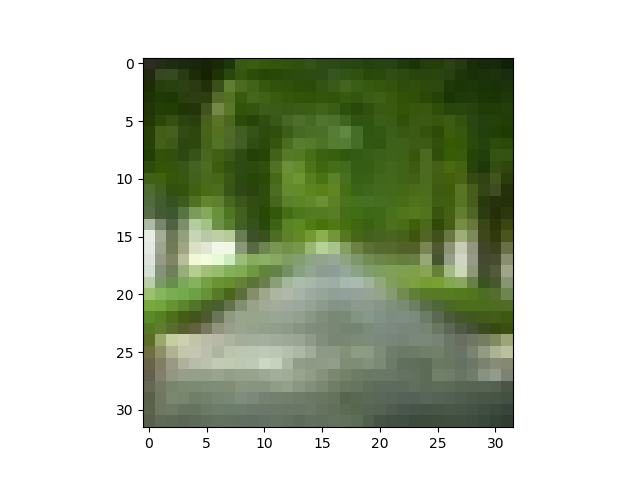}}
		\centerline{\includegraphics[width=\linewidth]{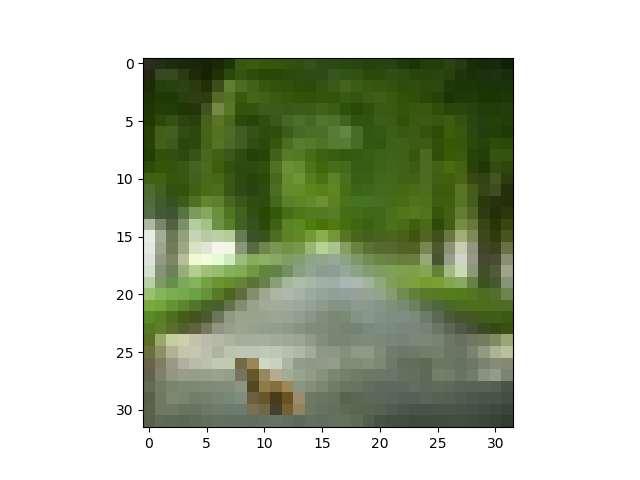}}
		\subcaption{marmot}\label{fig:Attack_F_exp}
	\end{minipage}
	\begin{minipage}[b]{0.25\linewidth}
		\centering
		\centerline{\includegraphics[width=\linewidth]{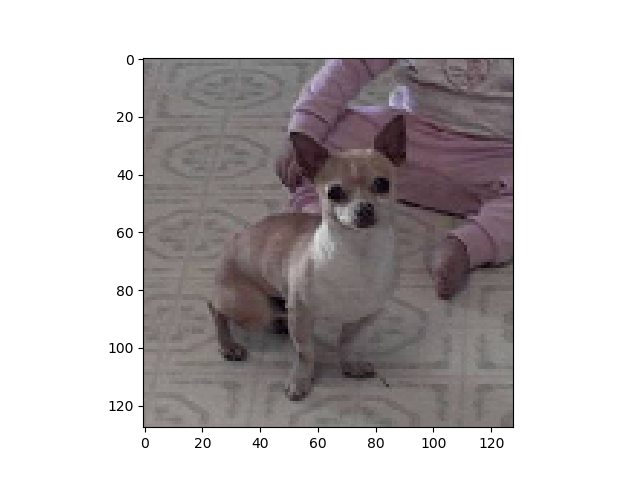}}
		\centerline{\includegraphics[width=\linewidth]{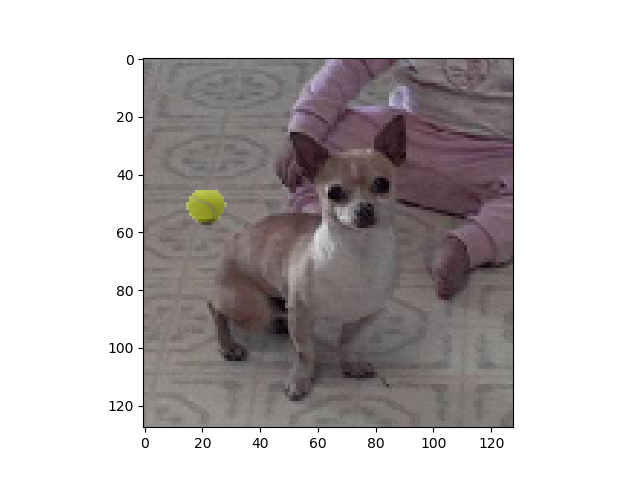}}
		\subcaption{tennis ball}\label{fig:Attack_G_exp}
	\end{minipage}
	\begin{minipage}[b]{0.25\linewidth}
		\centering
		\centerline{\includegraphics[width=\linewidth]{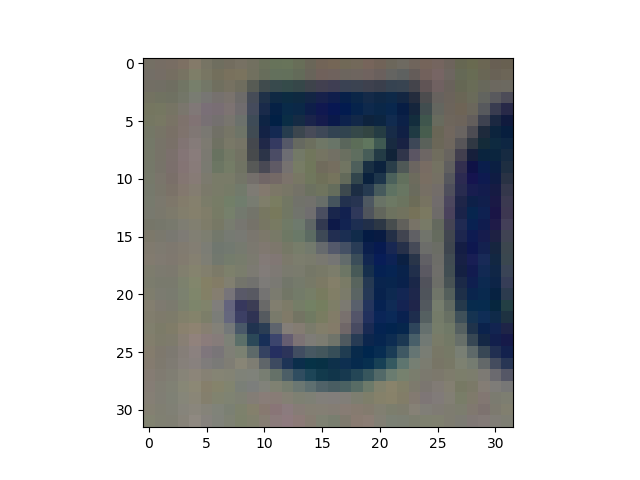}}
		\centerline{\includegraphics[width=\linewidth]{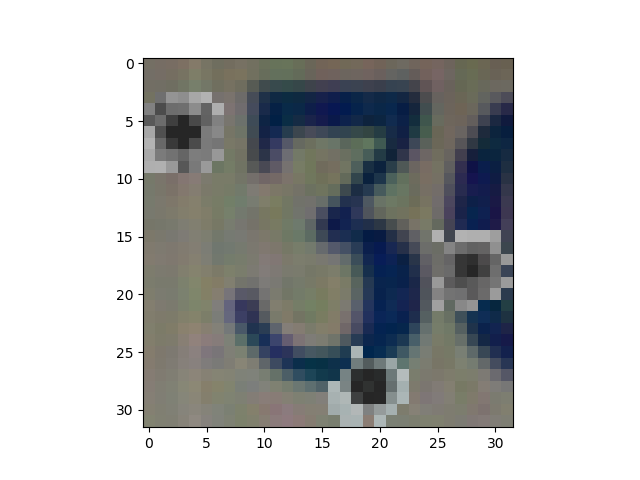}}
		\subcaption{bullet holes}\label{fig:Attack_H_exp}
	\end{minipage}
	\begin{minipage}[b]{0.25\linewidth}
		\centering
		\centerline{\includegraphics[width=\linewidth]{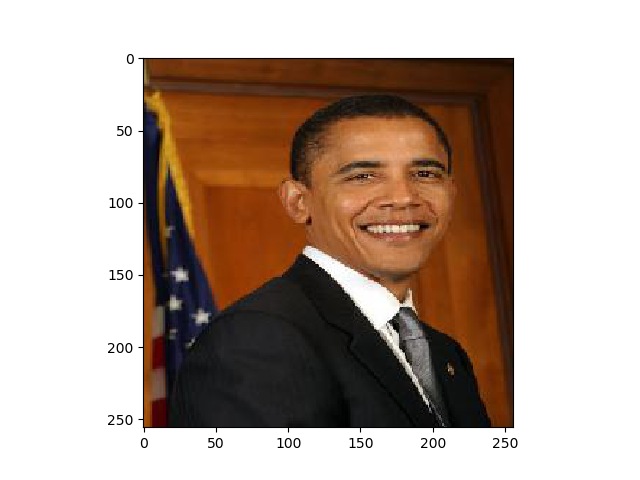}}
		\centerline{\includegraphics[width=\linewidth]{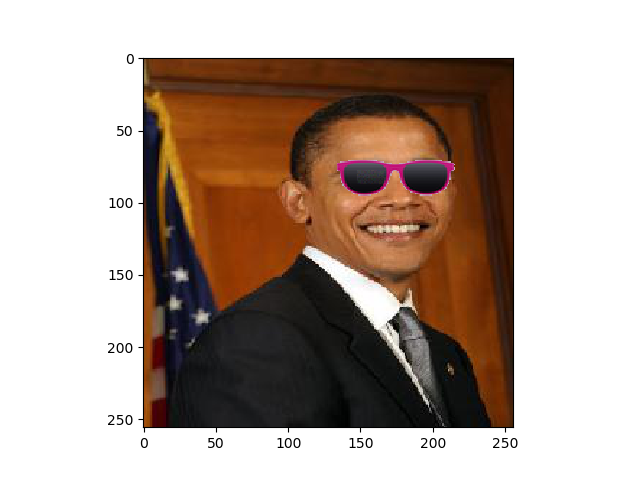}}
		\subcaption{sunglasses}\label{fig:Attack_I_exp}
	\end{minipage}
	\caption{Example backdoor image and the originally clean image for Attack A--I. Subcaptions describe the object(s) added as the perceptible backdoor pattern for each attack.}
	\label{fig:bd_examples_all}
\end{figure}

\begin{figure}[t]
	\centering
	\begin{minipage}[b]{.3\linewidth}
		\centering
		\centerline{\includegraphics[width=\linewidth]{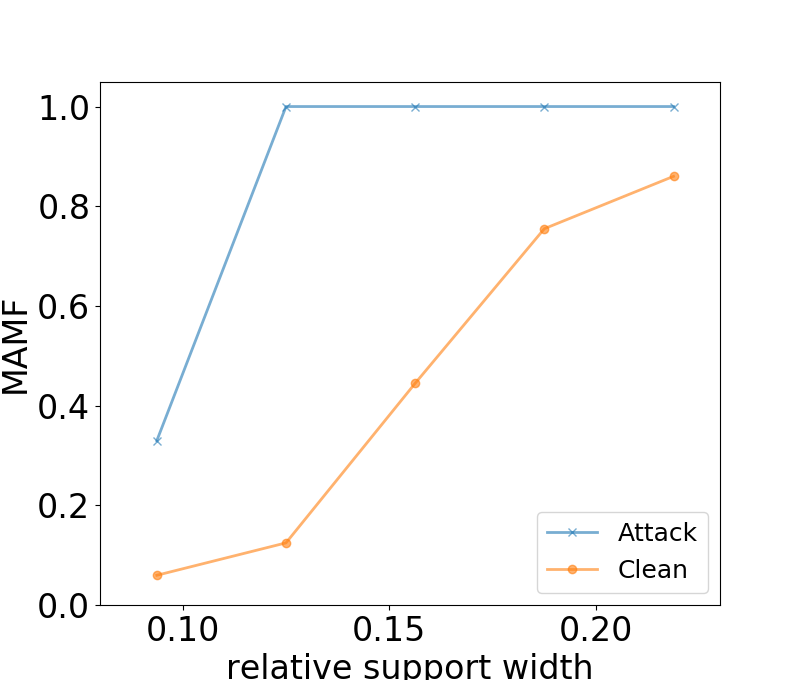}}
		\subcaption{Attack A}\label{fig:Attack_A_rho}
	\end{minipage}
	\begin{minipage}[b]{.3\linewidth}
		\centering
		\centerline{\includegraphics[width=\linewidth]{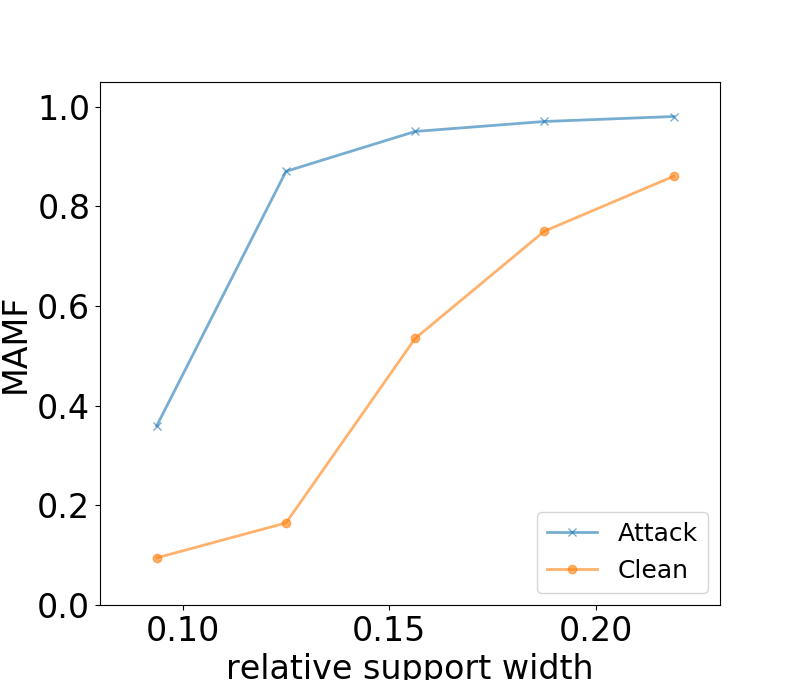}}
		\subcaption{Attack B}\label{fig:Attack_B_rho}
	\end{minipage}
	\begin{minipage}[b]{.3\linewidth}
		\centering
		\centerline{\includegraphics[width=\linewidth]{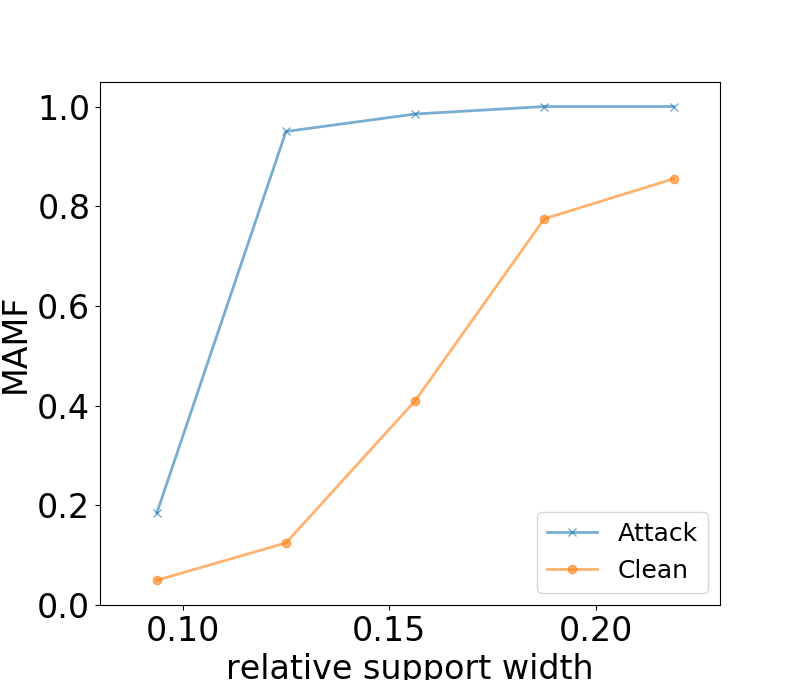}}
		\subcaption{Attack C}\label{fig:Attack_C_rho}
	\end{minipage}
	\begin{minipage}[b]{.3\linewidth}
		\centering
		\centerline{\includegraphics[width=\linewidth]{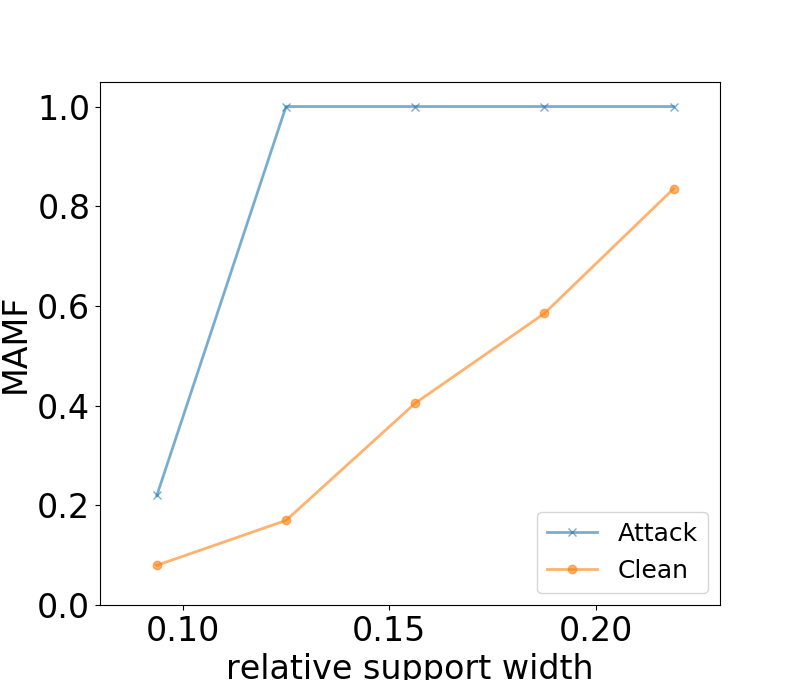}}
		\subcaption{Attack D}\label{fig:Attack_D_rho}
	\end{minipage}
	\begin{minipage}[b]{.3\linewidth}
		\centering
		\centerline{\includegraphics[width=\linewidth]{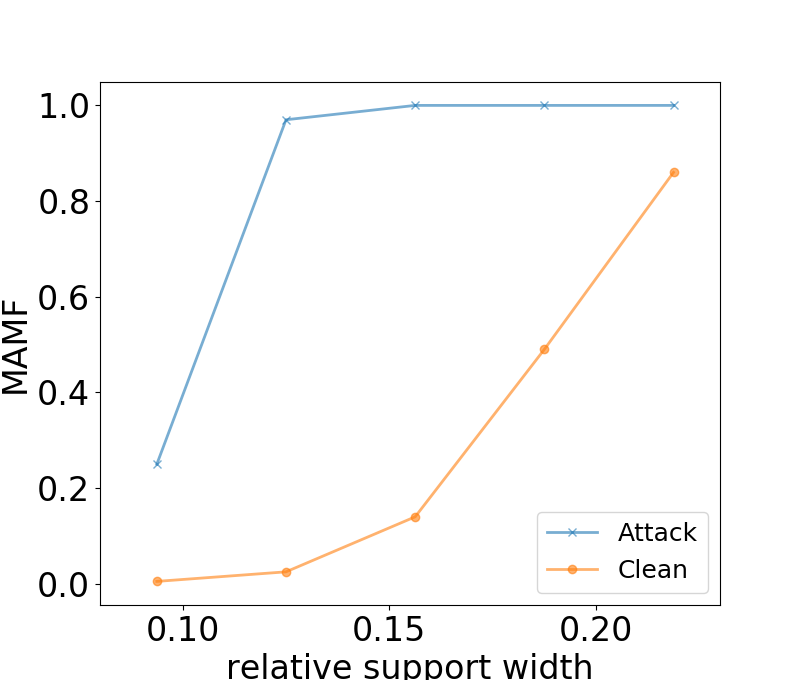}}
		\subcaption{Attack E}\label{fig:Attack_E_rho}
	\end{minipage}
	\begin{minipage}[b]{.3\linewidth}
		\centering
		\centerline{\includegraphics[width=\linewidth]{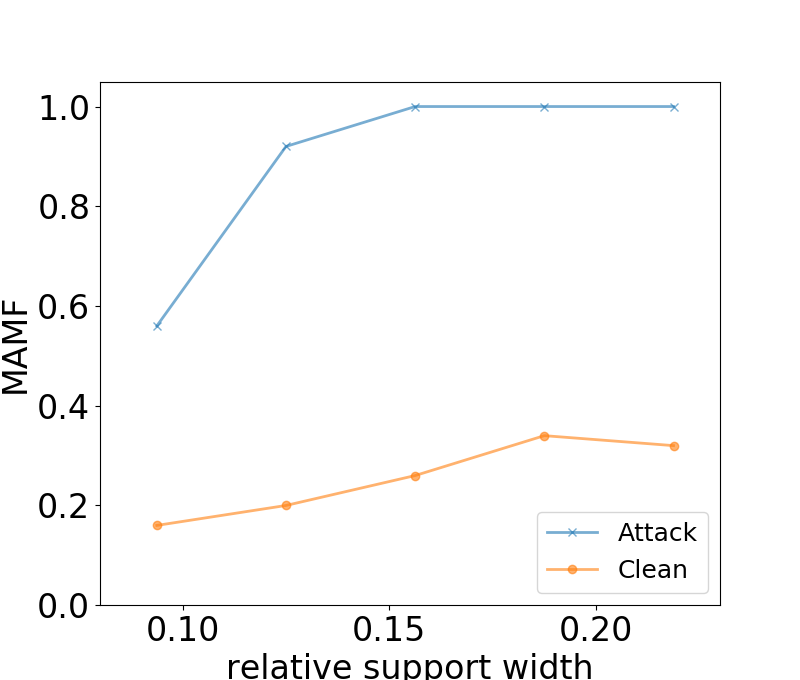}}
		\subcaption{Attack F}\label{fig:Attack_F_rho}
	\end{minipage}
	\begin{minipage}[b]{.3\linewidth}
		\centering
		\centerline{\includegraphics[width=\linewidth]{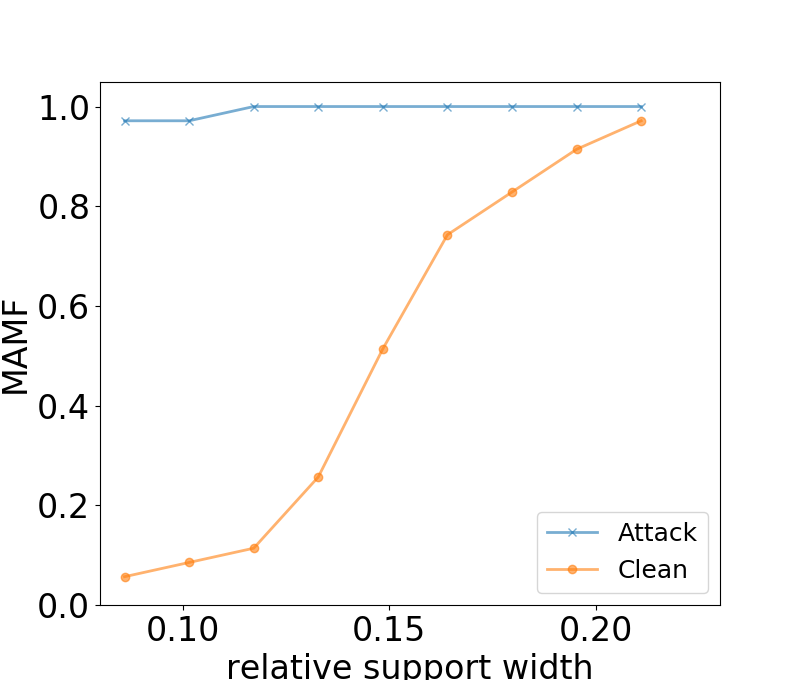}}
		\subcaption{Attack G}\label{fig:Attack_G_rho}
	\end{minipage}
	\begin{minipage}[b]{.3\linewidth}
		\centering
		\centerline{\includegraphics[width=\linewidth]{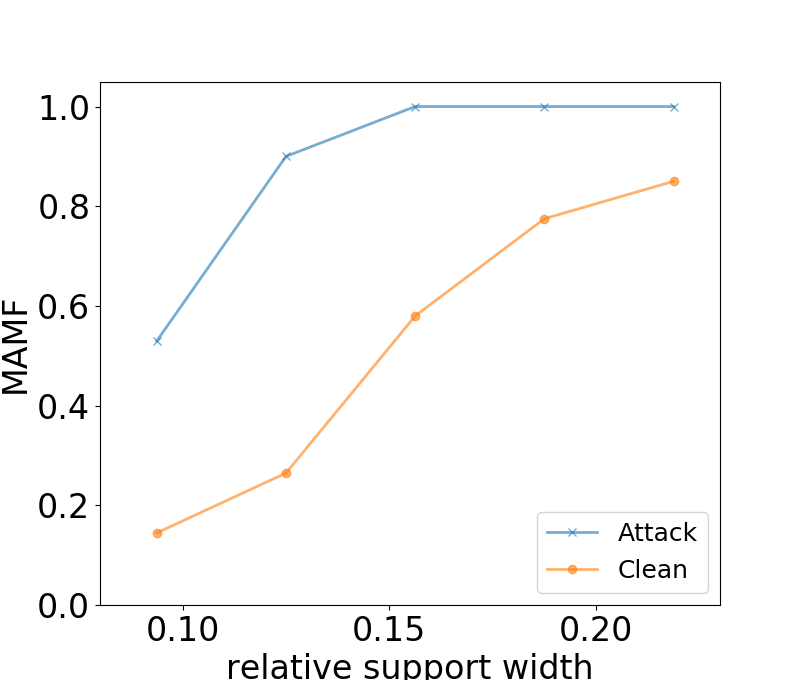}}
		\subcaption{Attack H}\label{fig:Attack_H_rho}
	\end{minipage}
	\begin{minipage}[b]{.3\linewidth}
		\centering
		\centerline{\includegraphics[width=\linewidth]{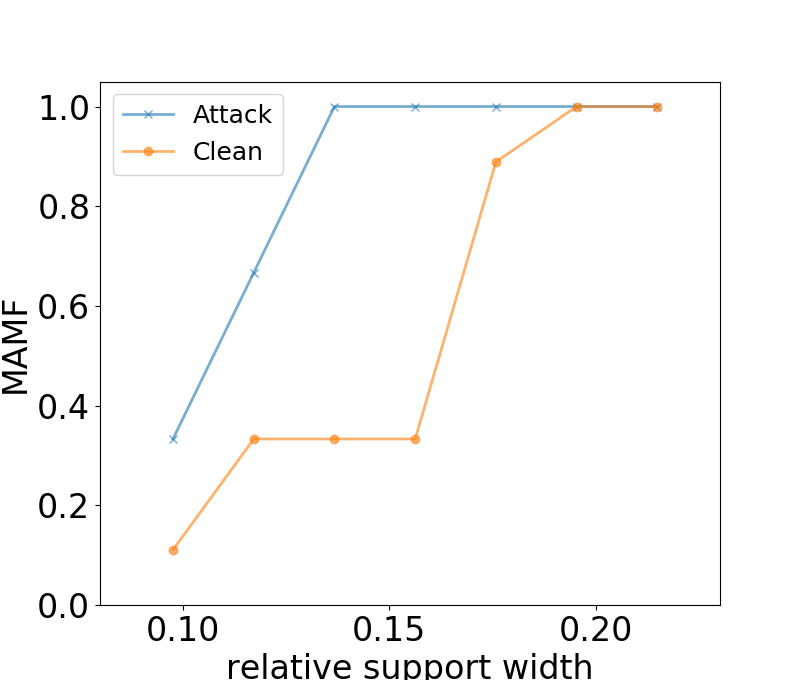}}
		\subcaption{Attack I}\label{fig:Attack_I_rho}
	\end{minipage}
	\caption{Maximum achievable misclassification fraction (MAMF) statistics for the class pair with the largest average MAMF, for both attacked DNN and unattacked (clean DNN), for Attack A--I. Relative support range for pattern estimation is $(0.08, 0.22)$.}
	\label{fig:rho_all}
\end{figure}

To create scene-plausible backdoor attacks, we first created a large number of candidate backdoor images, each with the backdoor pattern randomly located in the image. Then we manually picked the images in which the backdoor pattern looks scene-plausible\footnote{Only for Attack I we carefully place the sunglasses on the face.}. For example, for Attack C where the source class is ``airplane'' and the backdoor pattern is a ``rainbow'', a valid backdoor training image being selected should have the rainbow in the sky (see Figure \ref{fig:Attack_C_exp}). Since the focus of this paper is on the defense against attacks, we did not polish the backdoor patterns to make them completely innocuous to humans. In Figure \ref{fig:bd_examples_all}, we show an example backdoor training image and its original clean image for each attack. The backdoor patterns being considered here are representative of multiple types of practical backdoor patterns. The ``bug'' for Attack A and the ``butterfly'' for attack B represent backdoor patterns in the periphery (not covering the foreground object of interest) of an image and with a modest size. The ``rainbow'' for Attack C represents large backdoor patterns with an irregular shape. The backdoor patterns for Attack D and H represent dispersed patterns. The ``sunglasses'' for Attack I represent backdoor patterns overlapping with features of interest.

In Table \ref{tab:attacks_description}, we also report the accuracy on clean test images and the attack success rate for all attacks. The attack success rate is defined as the fraction of backdoor {\it test} images being classified to the target class prescribed by the attacker. A backdoor test image is created using a clean test image from the source class(es), adding to it the same backdoor pattern used for creating the backdoor training images. We took an automated approach (due to the huge number of test images) to create backdoor test images, randomly placing the backdoor pattern in the image (except for Attack I where we manually place the sunglasses on the faces). By doing so for a sufficiently large test set, we should be covering all possible spatial locations where the attacker could place the backdoor pattern when creating backdoor test images in practice. As shown in the table, for all attacks, the attack success rate is high, and there is no significant degradation in clean test accuracy compared with the benchmark accuracy; hence all attacks are considered successful. {\it Moreover}, we emphasize that such success is achieved with randomly placed backdoor patterns in test images, which experimentally verifies Property \ref{prop:spatial_invariance}, i.e. the spatial invariance of the learned perceptible backdoor mapping.

\subsection{Detection Performance Evaluation}\label{sec:performance_eval}

In this section, we evaluate the detection performance of the proposed detector, in comparison with the state-of-the-art detector NC \cite{NC}, and an even earlier defense FP \cite{FP}, using the above-mentioned attacks.

\begin{figure}[t]
	\centering
	\begin{minipage}[b]{0.7\linewidth}
		\centering
		\centerline{\includegraphics[width=\linewidth]{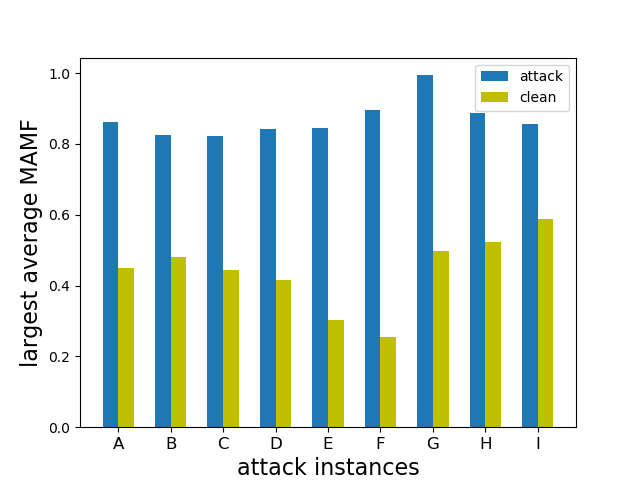}}
	\end{minipage}
	\caption{Largest average maximum achievable misclassification fraction (MAMF), i.e. $\rho^{\ast}$, over all class pairs for both the attacked DNN and unattacked (clean) DNN, for Attack A--I.}
	\label{fig:main_results}
\end{figure}

For each attack, detection is applied to both the DNN being attacked and the clean benchmark DNN. For the best discrimination between the two categories of DNNs, for our proposed method, relatively small spatial supports should be used for pattern estimation, so that a DNN being attacked can have a much larger $\rho^{\ast}$ than a clean DNN. On the one hand, the spatial support should be smaller than that of the foreground objects associated with the actual classes; otherwise, the actual target class object(s) could possibly be reverse engineered, causing high group misclassification even for a clean DNN. Moreover, if $r_{\rm max}$ needs to be chosen large to achieve high $\rho^{\ast}$ for a DNN that was backdoor-poisoned, then the backdoor is likely not innocuous, i.e. it is likely an {\it overly large} perceptible pattern and likely not scene-plausible. Here, $r_{\rm max}=0.22$ for all attack instances for our experiments. In practice, an $r_{\rm max}$ could be adaptively chosen for each dataset as the {\it critical} spatial support allowing a proportion of class pairs (e.g. 50\%) to achieve a modest MAMF (e.g. 0.4) during the pattern estimation step. On the other hand, the spatial support should be large enough for estimating any pattern related to the backdoor. Here, $r_{\rm min}=0.08$, such that for a $32\times 32$ image, there is at least $3\times 3$ spatial support for pattern estimation. 

The details of our detection settings are as follows. For Attack A--E and Attack H, 200 clean images that are correctly classified {\it per class} are used for detection. For Attack F, G and I, 50, 30 and 9 correctly classified clean images per class are used for detection, respectively. For all DNNs, class pairs and spatial supports, we solve (\ref{eq:opt_main}) using Adam optimizer and learning rate 0.5 for 100 epochs\footnote{Fewer epochs are actually needed for decent convergence in all our experiments.}. We use mini-batch size 32 for Attack A--E and H, 10 for Attack F and G and 3 for Attack I. The spatial support for pattern estimation, used in our detection, is a square covering the top left corner of each image. Due to Property \ref{prop:spatial_invariance} that has been verified in Section \ref{sec:attack_crafting}, other locations yield similar results, as will be illustrated in Section \ref{sec:window_loc}.

In Figure \ref{fig:rho_all}, for both the DNN being attacked and the clean benchmark DNN, for each attack, we show the ($L$) MAMF statistics for the class pair with the largest average MAMF (i.e. the class pair corresponding to $\rho^{\ast}$). For example, for Attack A where the data image size is $32\times 32$, the (absolute) support widths being considered are $w=3, 4, 5, 6, 7$. For large image size, instead of performing pattern estimation for each integer support width in the interval $[\ceil{r_{\rm min}\times W}, \floor{r_{\rm max}\times W}]$, we could efficiently downsample and perform pattern estimation for fewer support widths. Here, $L=9$ and $L=7$ support widths are considered for Attack G (see Figure \ref{fig:Attack_G_rho}) and Attack I (see Figure \ref{fig:Attack_I_rho}) respectively. In each sub-figure of Figure \ref{fig:rho_all}, a large gap is observed between the clean and attacked curves, clearly distinguishing the DNN being attacked from the clean DNN.

In Figure \ref{fig:main_results}, we show the largest average MAMF (i.e. $\rho^{\ast}$) for both the clean and attacked DNNs, for all attacks. Note the clear large difference between the two bars for each attack. Thus any detection threshold $\pi$ in $(0.6, 0.8)$, if used, could successfully detect whether the DNN is attacked. Moreover, for the DNN being attacked, for all attack instances, the class pair corresponding to the maximum average MAMF is precisely the {\it true backdoor pair}. Hence when a detection is made, the (source, target) class pair (used by the attacker) is also correctly inferred, in all cases.

For the attacks in the current experiment (that use a single source class), NC is not expected to make a correct detection, since NC is based on the assumption that {\it all classes} other than the target class are involved in the attack. Also, NC jointly estimates a pattern and an associated mask for each putative target class to induce at least $\phi$-level misclassification. Such optimization relies on the choice of the penalty multiplier $\lambda$ (for the ${\rm L}_1$-regularization of the mask) and the training settings. Here we evaluate NC using the same attacks. We used the Adam optimizer with the parameters suggested by the authors of NC and performed mini-batch optimization for a sufficient number of epochs (until convergence). If $\lambda$ is too large, $\phi$-level group misclassification cannot be achieved since the mask ``size'' is over-penalized. If $\lambda$ is made small, high group misclassification fraction will be maintained, but the ${\rm L}_1$ norm of the mask is unreasonably large. Hence we carefully adjust $\lambda$ and the training parameters for each attack to achieve a mask with small ${\rm L}_1$ norm and a pattern inducing a high group misclassification fraction to the true backdoor class.  In this way, we in fact are optimistically tuning NC's hyperparameter $\lambda$ to maximize NC's performance. 
For all the attack cases, we set $\phi=0.9$, and the optimization is performed for 200 epochs. The number of clean images per class used for pattern estimation in the NC detection procedure, the choice of $\lambda$, the learning rate and the batch size for each attack instance are shown in Table \ref{tab:NC_details}.

\begin{table*}[t]
	\begin{center}
		\caption{Detailed settings of NC, including the number of clean images per class used for detection, the choice of $\lambda$, the learning rate and the batch size, for each attack instance.}
		\resizebox{1.0\textwidth}{!}{
			\begin{tabular}{ |c|c|c|c|c|c|c|c|c|c|c| } 
				\hline
				& Attack A & Attack B & Attack C & Attack D & Attack E & Attack F & Attack G & Attack H & Attack I\\ 
				\hline
				No. images per class & 100 & 100 & 100 & 100 & 100 & 10 & 30 & 100 & 9\\
				\hline
				$\lambda$ & 0.1 & 0.1 & 0.1 & 0.1 & 0.6 & 0.1 & 0.5 & 0.2 & 0.1\\
				\hline
				Learning rate & 0.05 & 0.05 & 0.05 & 0.05 & 0.05 & 0.05 & 0.001 & 0.01 & 0.005\\
				\hline
				Batch size & 90 & 90 & 90 & 90 & 90 & 90 & 60 & 100 & 30\\
				\hline
			\end{tabular}\label{tab:NC_details}}
	\end{center}
\end{table*}

\begin{figure}[t]
	\centering
	\begin{minipage}[b]{0.7\linewidth}
		\centering
		\centerline{\includegraphics[width=\linewidth]{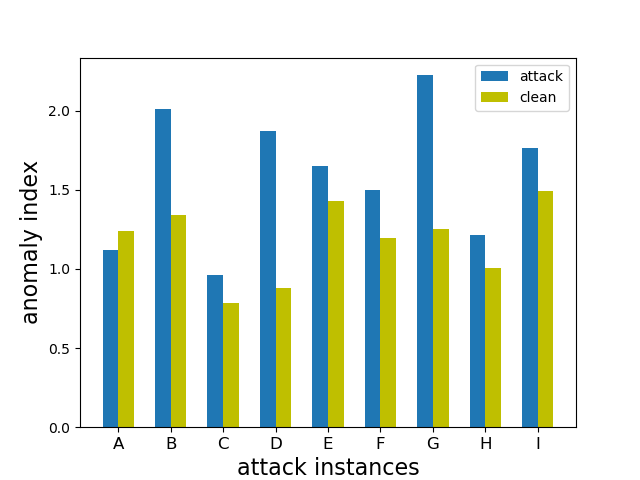}}
	\end{minipage}
	\caption{Anomaly indices when applying NC to the DNN being attacked and the clean DNN of Attack A--I.}
	\label{fig:NC_results}
\end{figure}

The detection inference of NC uses an ``anomaly index'' derived based on median absolute deviation (MAD) \cite{MAD} of the ${\rm L}_1$ norm of the mask for all putative target classes \cite{NC}. If the anomaly index is larger than 2.0, a detection is made with $95\%$ confidence. In Figure \ref{fig:NC_results}, we show the anomaly indices for both the DNN being attacked and the clean DNN for each attack instance. Only for Attack B and Attack G, NC successfully detects the attack -- for these two attacks, there is a pattern and an associated {\it relatively small} mask which, when applied to clean test images from all classes other than the target class, induces a high group misclassification (even though the backdoor targeted only a single source class). This phenomenon (where non-source class images are with high probability misclassified to the target class when the backdoor pattern is added to them) was discovered and identified in \cite{Post-TNNLS} as ``collateral damage'' (ref. Section \ref{sec:source_class}).

\begin{figure}[t]
	\centering
	\begin{minipage}[b]{.3\linewidth}
		\centering
		\centerline{\includegraphics[width=\linewidth]{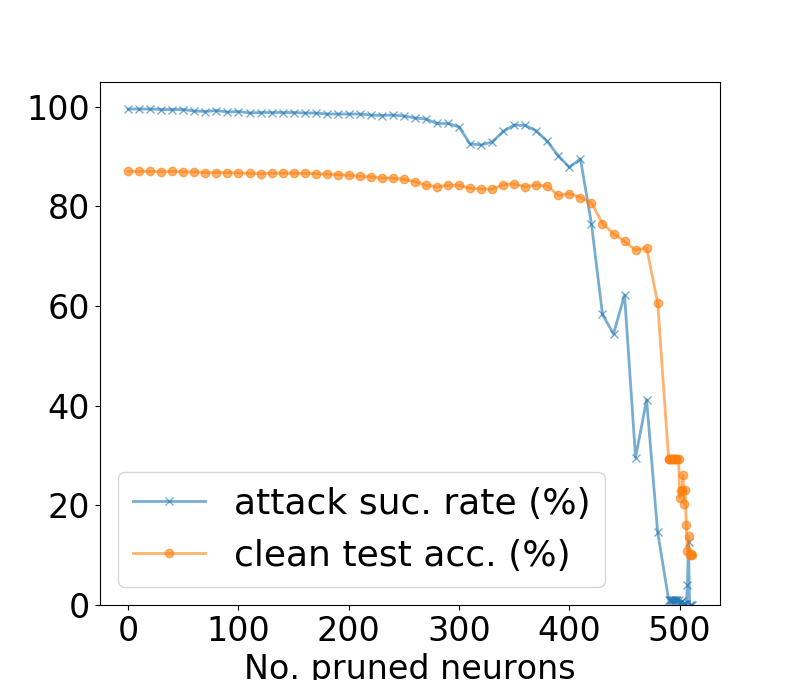}}
		\subcaption{Attack A}\label{fig:FP_A}
	\end{minipage}
	\begin{minipage}[b]{.3\linewidth}
		\centering
		\centerline{\includegraphics[width=\linewidth]{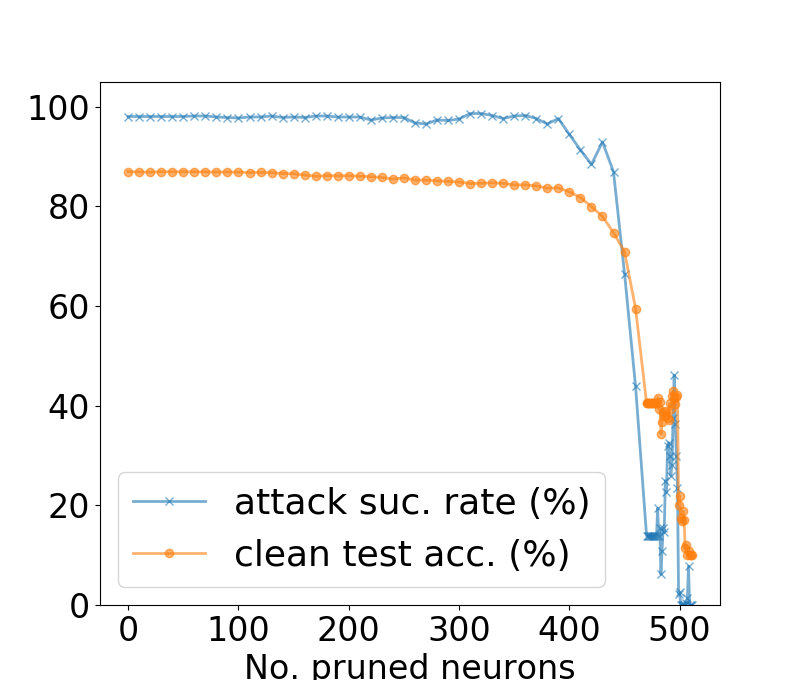}}
		\subcaption{Attack B}\label{fig:FP_B}
	\end{minipage}
	\begin{minipage}[b]{.3\linewidth}
		\centering
		\centerline{\includegraphics[width=\linewidth]{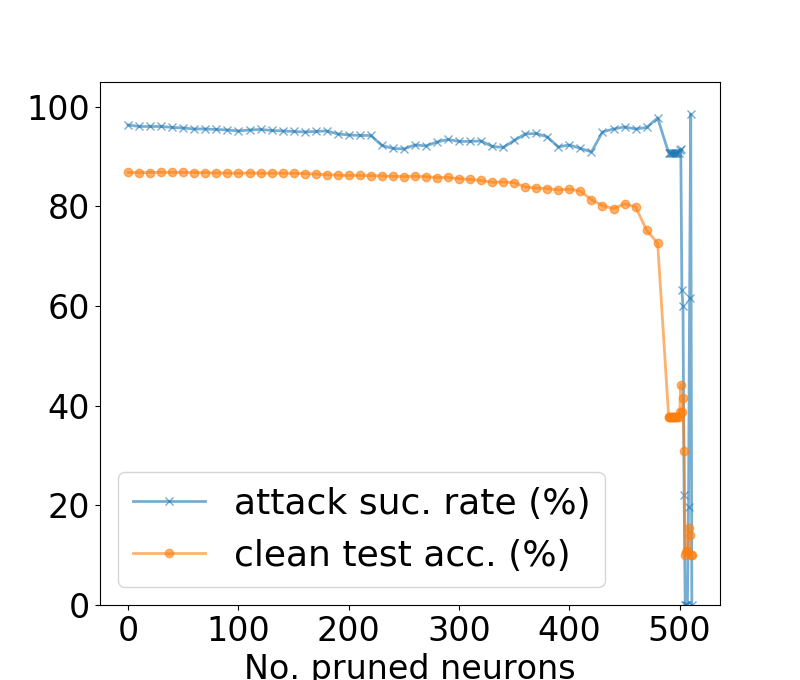}}
		\subcaption{Attack C}\label{fig:FP_C}
	\end{minipage}
	\begin{minipage}[b]{.3\linewidth}
		\centering
		\centerline{\includegraphics[width=\linewidth]{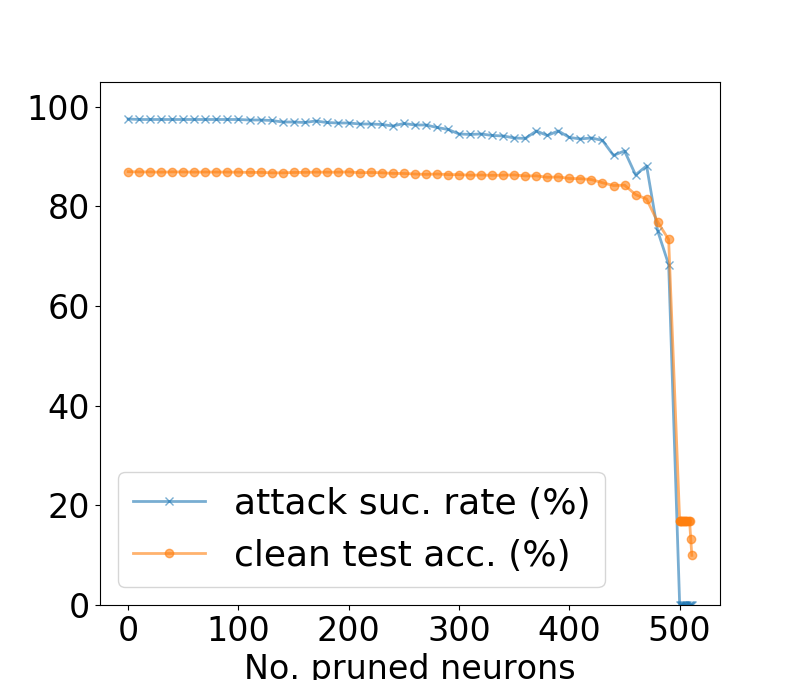}}
		\subcaption{Attack D}\label{fig:FP_D}
	\end{minipage}
	\begin{minipage}[b]{.3\linewidth}
		\centering
		\centerline{\includegraphics[width=\linewidth]{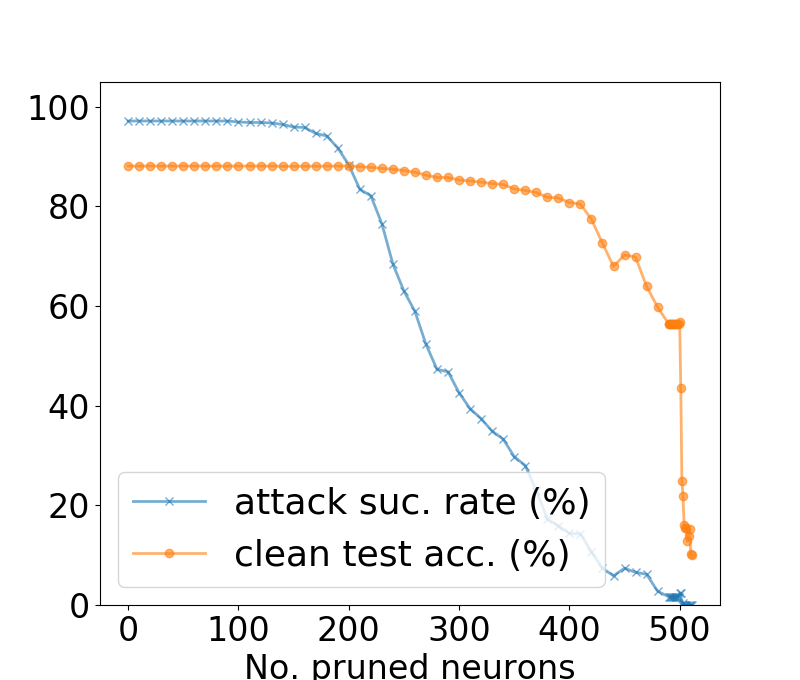}}
		\subcaption{Attack E}\label{fig:FP_E}
	\end{minipage}
	\begin{minipage}[b]{.3\linewidth}
		\centering
		\centerline{\includegraphics[width=\linewidth]{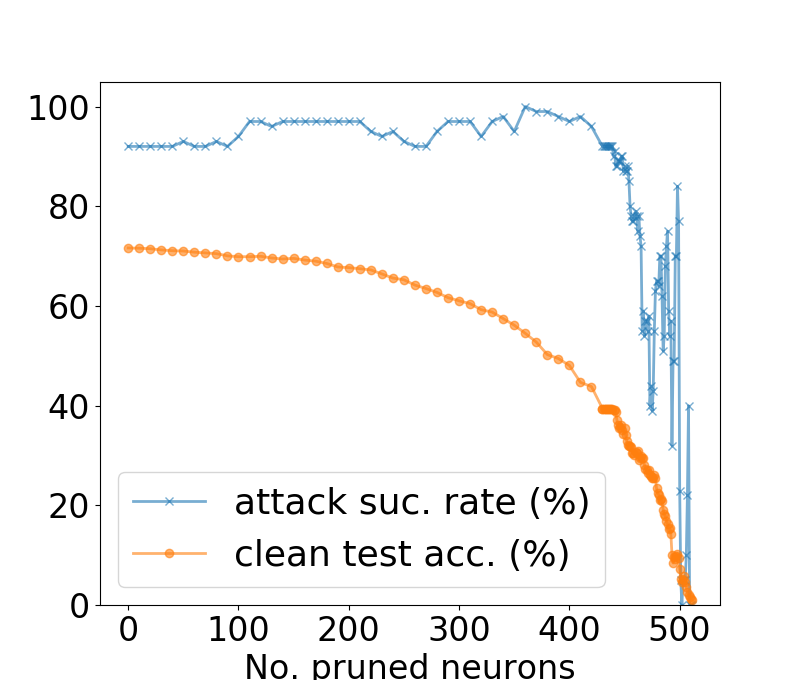}}
		\subcaption{Attack F}\label{fig:FP_F}
	\end{minipage}
	\begin{minipage}[b]{.3\linewidth}
		\centering
		\centerline{\includegraphics[width=\linewidth]{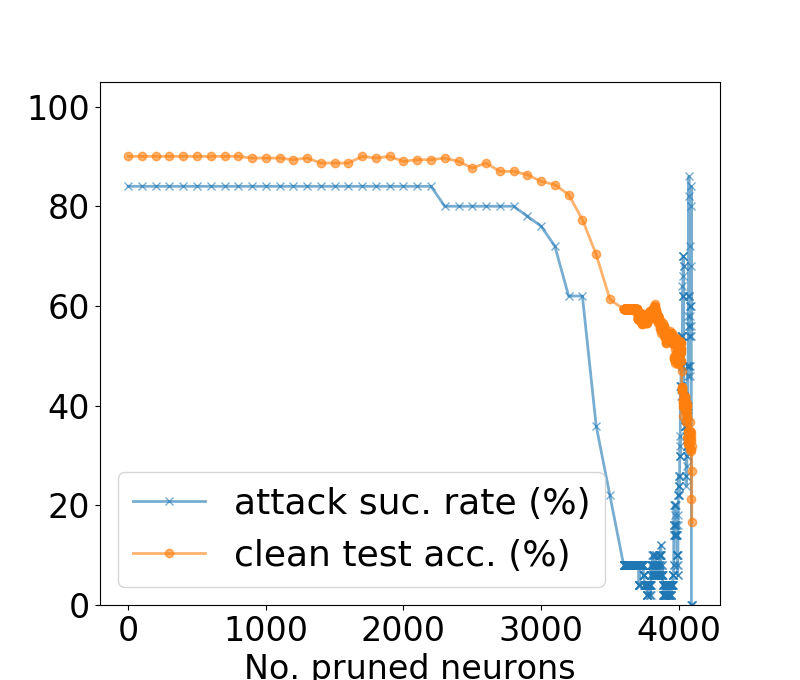}}
		\subcaption{Attack G}\label{fig:FP_G}
	\end{minipage}
	\begin{minipage}[b]{.3\linewidth}
		\centering
		\centerline{\includegraphics[width=\linewidth]{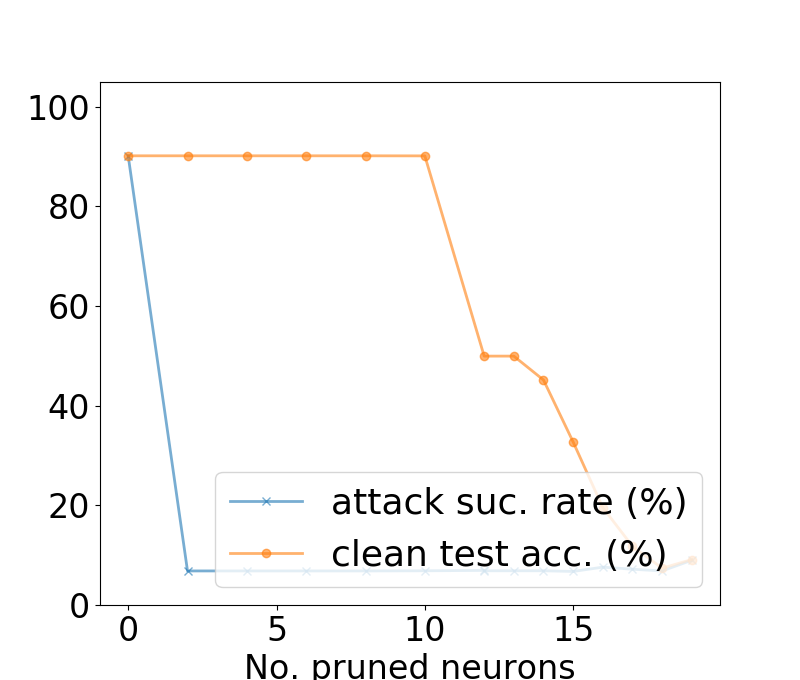}}
		\subcaption{Attack H}\label{fig:FP_H}
	\end{minipage}
	\begin{minipage}[b]{.3\linewidth}
		\centering
		\centerline{\includegraphics[width=\linewidth]{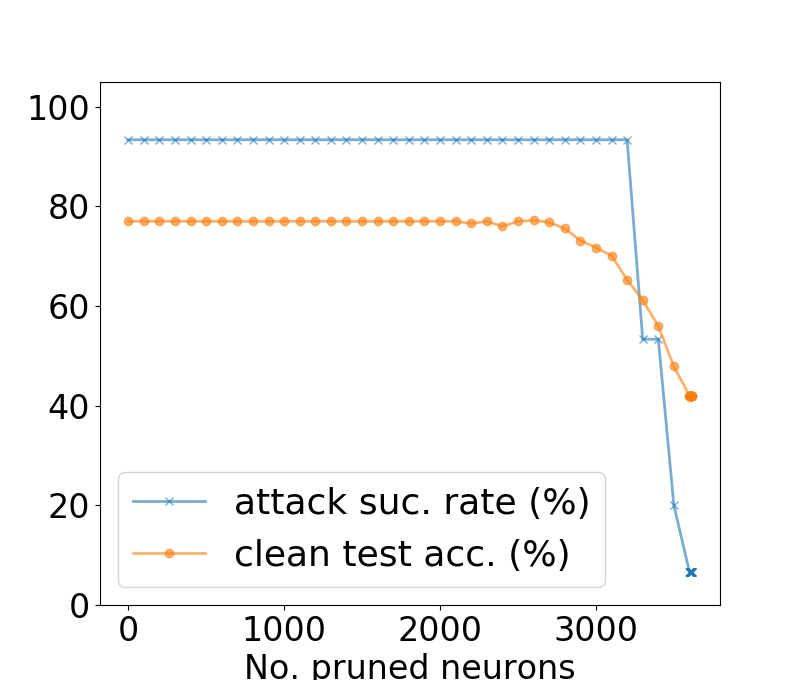}}
		\subcaption{Attack I}\label{fig:FP_I}
	\end{minipage}
	\caption{Attack success rate and accuracy on clean test images as the number of penultimate layer neurons being pruned is increased, for each DNN being attacked.}
	\label{fig:FP_all}
\end{figure}
We also show that the FP defense proposed by \cite{FP} is ineffective against most of our devised attacks. We prune the penultimate layer neurons of each classifier being attacked (until only few are left), in increasing order of their average activations over all clean test images. In Figure \ref{fig:FP_all}, for each attack, we show the accuracy on clean test images and the attack success rate versus the number of neurons being pruned. For most of the attacks (except Attack E and H), the attack success rate does not drop before the accuracy on clean test images is degraded, as the number of neurons being pruned grows.
Thus, the FP defense is generally unsuccessful against these backdoor attacks.
\subsection{Verification of Property 2}\label{sec:verify_prop_2}

\begin{table}[t]
	\begin{center}
		\caption{Attack success rate (\%) of backdoor test images with (Gaussian) noisy backdoor patterns with $\sigma^2=0.01, 0.25, 1$ and cropped backdoor patterns to 64\% and 36\% of the original size.}
		\resizebox{0.36\textwidth}{!}{
			\begin{tabular}{ |c|c|c|c|c|c| } 
				\hline
				& \multicolumn{3}{|c|}{$\sigma^2$} & \multicolumn{2}{|c|}{Crop}\\
				\hline
				& 0.01 & 0.25 & 1 & 64\% & 36\%\\ 
				\hline
				Attack A & 85.2 & 53.6 & 53.7 & 84.6 & 73.2\\
				\hline
				Attack B & 97.8 & 86.7 & 87.6 & 67.8 & 26.3\\
				\hline
				Attack C & 62.4 & 46.9 & 23.6 & 96.0 & 29.1\\
				\hline
				Attack D & 98.1 & 99.9 & 99.0 & 81.7 & 60.2\\
				\hline
				Attack E & 78.4 & 45.6 & 32.0 & 91.4 & 46.0\\
				\hline
				Attack F & 97.0 & 91.0 & 83.0 & 86.0 & 69.0\\
				\hline
				Attack G & 78.0 & 38.0 & 28.0 & 78.0 & 62.0\\
				\hline
				Attack H & 91.3 & 66.7 & 67.2 & 41.7 & 20.7\\
				\hline
				Attack I & 86.7 & 86.7 & 80.0 & 40.0 & 26.7\\
				\hline
			\end{tabular}\label{tab:verify_prop_2}}
	\end{center}
\end{table}

Here we verify Property \ref{prop:robustness}, the robustness property of perceptible backdoor patterns, from two aspects. First, for each attack, we modify the backdoor pattern embedded into {\it each} clean test image (from the source class) by adding Gaussian noise $N(0, \sigma^2)$ to each pixel (and then clipping each pixel value to [0, 1]). Second, instead of adding noise, we crop part of the backdoor pattern outside its center before embedding to the clean test images. We evaluate the attack success rate for noisy backdoor patterns with $\sigma^2=0.1, 0.5, 1$ and cropped backdoor patterns to 64\% and 36\% of the original size in Table \ref{tab:verify_prop_2}. For most of the attack instances, using noisy or cropped backdoor patterns at test time still achieves high attack success rate (even though cropping around the center of a backdoor pattern may remove critical features on the periphery of the backdoor pattern).

\subsection{Number of Images for Detection}\label{sec:NI}

\begin{figure}[t]
	\centering
	\begin{minipage}[b]{0.95\linewidth}
		\centering
		\centerline{\includegraphics[width=\linewidth]{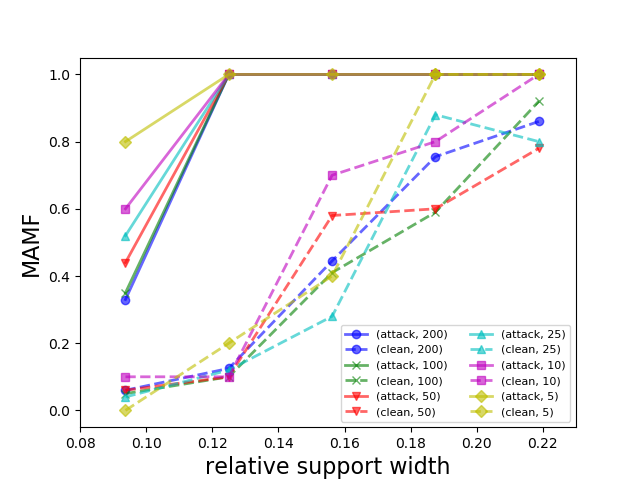}}
	\end{minipage}
	\caption{Maximum achievable misclassification fraction (MAMF) $\rho^{\ast}$ for a range of choices of relative support width $r$, for both DNNs being attacked and unattacked for Attack A, when using 5, 10, 25, 50, 100 and 200 clean images per class for detection, respectively.}
	\label{fig:NI}
\end{figure}

We note that if too few clean images are available to the defender, the performance of our detector could be affected. When there are no attacks, a false detection may be made since it is much easier to find a pattern that induces a high misclassification rate for a small group, compared with a larger one. This has been essentially verified in \cite{DeepFool_Univ}, as a ``universal'' TTE perturbation (a common TTE perturbation applied to all images) inducing high group misclassification needs much larger norm (and full spatial support) than a perturbation inducing single-image misclassification at test time. In Figure \ref{fig:NI}, we show the MAMF statistics for the class pair with the largest average MAMF for the clean and attacked DNNs for Attack A, with 5, 10, 25, 50, 100 and 200 clean images per class used for pattern estimation. When the number of clean images used for detection is greater than 10, there is a clear gap between the two curves for the DNN being attacked and its clean counterpart. But if fewer images are used for detection, the curve corresponding to the clean DNN approaches 1.0 quite quickly, i.e. achieving a high group misclassification fraction becomes easier and requires a smaller spatial support. This is the same phenomenon as in Figure \ref{fig:Attack_I_rho} for Attack I, where the curve for the clean DNN quickly achieves 1.0 as the support width for pattern estimation increases, since only 9 clean images are used for detection. Note, though, that our detector is successful even in this case -- there is still a large gap between the curves.

In practice, for the purpose of evaluating the clean test accuracy of the DNN (clean test images can be used both in a backdoor detection procedure and for evaluating clean test accuracy), the user should possess sufficient clean images from all classes. If not, we suggest to use a smaller detection threshold to compensate.

\subsection{Backdoor Patterns with Fixed Spatial Location}\label{sec:fixed_loc}

Here we perform a simple experiment to show that the power of the {\it attack} will be degraded if the perceptible backdoor pattern is spatially fixed when poisoning the clean training images. We use the same dataset, training settings, (source, target) class pair and backdoor pattern under Attack B. The only difference is that, the backdoor training images are created by embedding the backdoor pattern, the ``butterfly'', into the bottom left corner of {\it every} clean training image to be poisoned.

After training, the accuracy of the DNN on clean test images is 87.2\%, which is similar to the accuracy of the clean benchmark DNN. Now we create four groups of images, 1000 each, using the clean test images from the source class prescribed by the attacker, such that the backdoor pattern is embedded:

1) into the bottom left corner (i.e. the same location as for the backdoor training images) of all images.

2) spatially randomly in each image.

3) one row up from the bottom left corner of all images.

4) one column right of the bottom left corner of all images.

Here we only focus on the spatial location of the backdoor pattern without considering whether the backdoor patterns are scene-plausible or not. The fraction (\%) of images in each group that are (mis)classified to the target class prescribed by the attack are 99.7, 4.1, 47.8, 14.4, respectively. Clearly, only if the backdoor pattern during testing is located at the {\it same} place as during training, the backdoor image will be reliably (mis)classified to the target class. Hence fixing the spatial location of the backdoor pattern when creating backdoor training images largely degrades the robustness of the attack.

\subsection{Location of the Spatial Support for Pattern Estimation}\label{sec:window_loc}

According to Property \ref{prop:spatial_invariance}, the spatial invariance of the perceptible backdoor mapping, the location of the spatial support in the pattern estimation step can be arbitrarily chosen. Previously, we fixed the spatial support location to cover the top left corner for all clean images used for detection. Here, we experimentally verify that such choice is not critical to our detection.

\begin{figure}
	\centering
	\begin{minipage}[b]{.3\linewidth}
		\centering
		\centerline{\includegraphics[width=\linewidth]{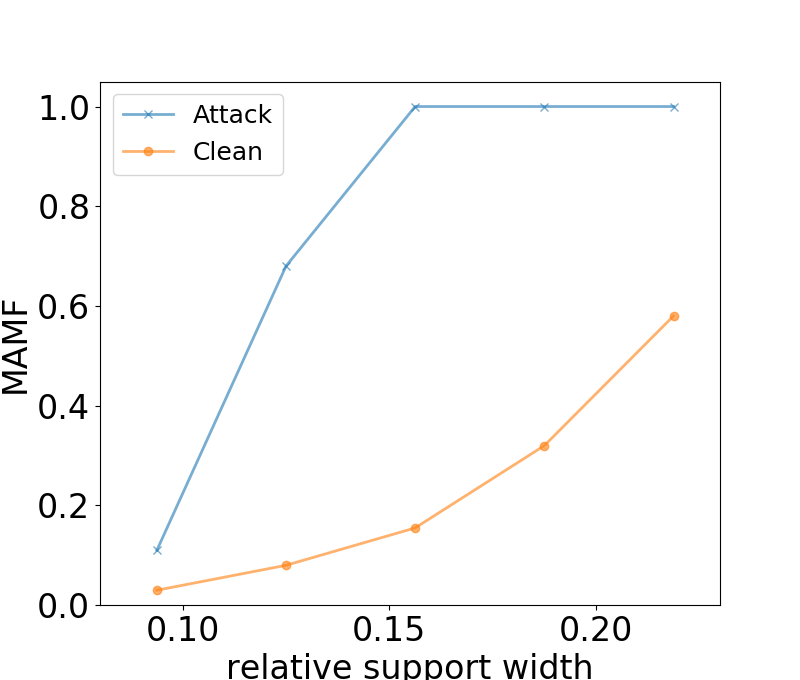}}
		\subcaption{Top right}\label{fig:corner_tr}
	\end{minipage}
	\begin{minipage}[b]{.3\linewidth}
		\centering
		\centerline{\includegraphics[width=\linewidth]{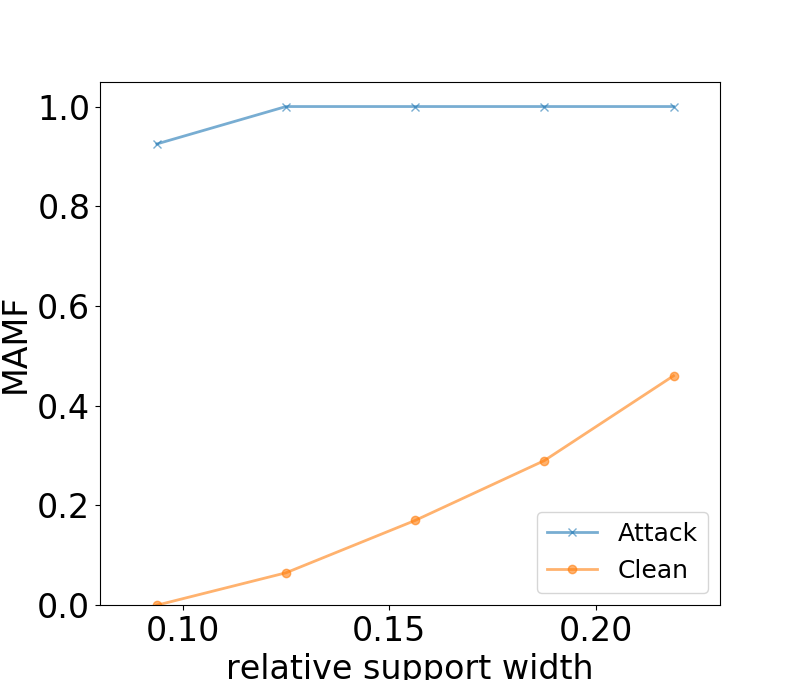}}
		\subcaption{Bottom left}\label{fig:corner_bl}
	\end{minipage}
	\begin{minipage}[b]{.3\linewidth}
		\centering
		\centerline{\includegraphics[width=\linewidth]{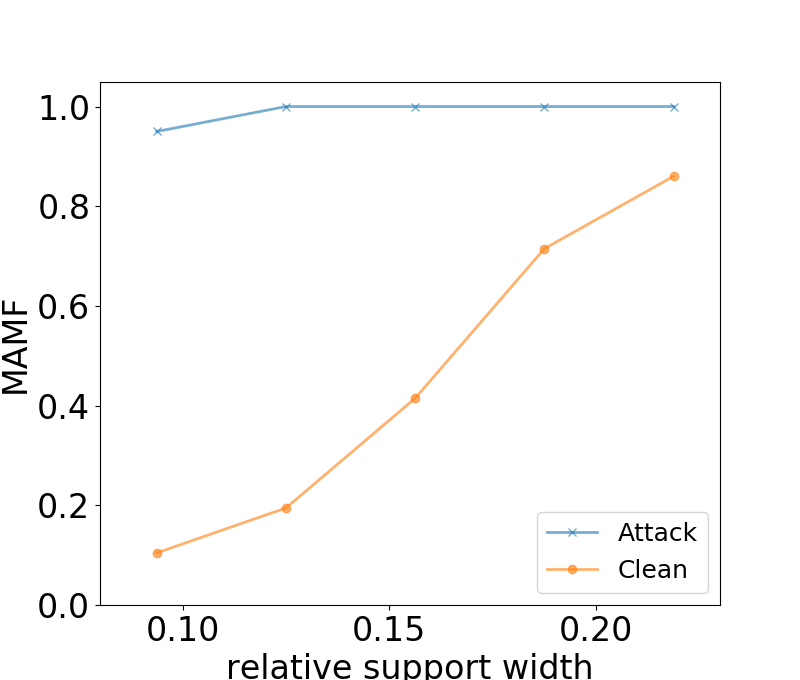}}
		\subcaption{Bottom right}\label{fig:corner_br}
	\end{minipage}
	\caption{Maximum achievable misclassification fraction (MAMF) statistics for the class pair with the largest average MAMF, for both the DNN being attacked and the DNN not attacked under Attack B. The spatial support for pattern estimation is fixed to cover a) the top left corner, b) the bottom left corner, and c) the bottom right corner.}
	\label{fig:rho_corners}
\end{figure}

Again, we consider the same DNN being attacked and the same clean benchmark DNN under Attack B. We apply the same detection to both the clean and attacked DNNs except that the spatial support for detection is fixed to cover 1) the top right corner, 2) the bottom left corner, 3) the bottom right corner of all clean images used for detection. In Figure \ref{fig:rho_corners}, for each location of the spatial support, we show the $L$ MAMF statistics (by varying the support width) for the class pair with the largest average MAMF, for both DNNs. In each figure, we observe a large gap between the two curves, indicating that the DNN being attacked could be easily detected for a large range of thresholds. Thus, our detection approach is indeed robust to the chosen location of the spatial support of the pattern mask.

\subsection{Detecting Backdoor Attacks with Multiple Source Classes}\label{sec:multiple_source}

\begin{figure}[t]
	\centering
	\begin{minipage}[b]{0.8\linewidth}
		\centering
		\centerline{\includegraphics[width=\linewidth]{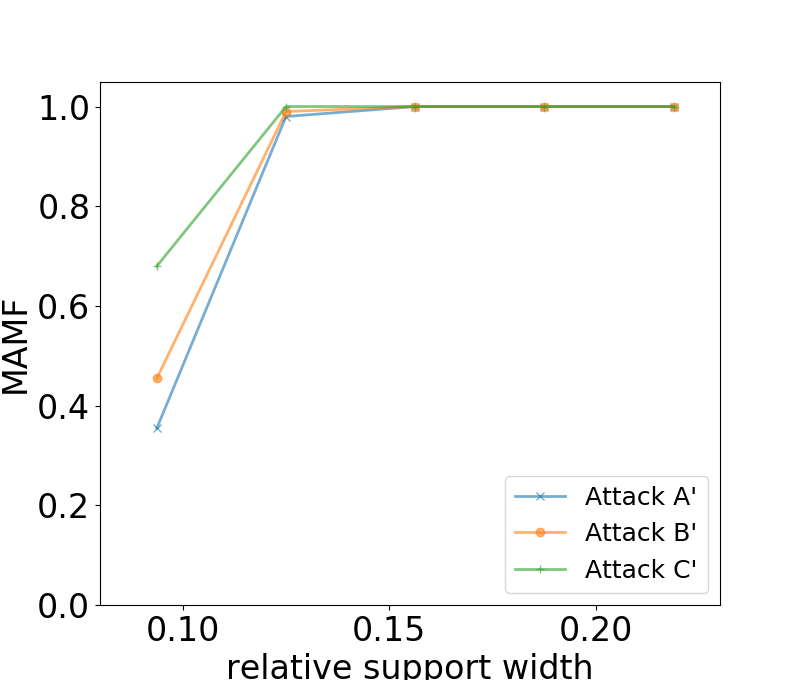}}
	\end{minipage}
	\caption{Maximum achievable misclassification fraction (MAMF) $\rho^{\ast}$ for a range of choices of relative support width $r$, for Attack A', B', and C'.}
	\label{fig:multi_source}
\end{figure}

Here, we show the performance of our detector against backdoor attacks involving multiple source classes. In particular, we consider the case where all classes other than the target class are the source classes. We devise Attack A', B', and C', using the same settings as Attack A, B, and C, respectively, but with all classes except the target class involved as the source classes and 20 backdoor training images per source class (i.e. 180 backdoor training images in total). Note that the purpose of this experiment is to show that our detector does not rely on knowledge of the number of source classes involved in an attack. We do not require the backdoor patterns used in this attack experiment to be scene-plausible. For example, for Attack C', we simply embed the backdoor pattern, the rainbow, into random locations of images from classes like ``frog'', ``dog'', and ``cat'', while most of images from these classes do not contain the sky. The attack success rate (\%) of Attack A', B', and C' are 98.0, 95.3, and 98.7, respectively; the accuracy (\%) on clean test images are 87.4, 86.8, 87.0, respectively. We apply our detector with the same setting as in Section \ref{sec:performance_eval} to these three attacks respectively and show the MAMF statistics for the class pair with the largest average MAMF, for each attack, in Figure \ref{fig:multi_source}. Comparing with the benchmark MAMF statistics shown in Figure \ref{fig:Attack_A_rho}, \ref{fig:Attack_B_rho}, and \ref{fig:Attack_C_rho}, the three attacks (with multiple source classes) could be easily detected by our detector.

\subsection{Detecting Perceptible Backdoors In-Flight}\label{sec:TTE_in_flight}

Here we show that the TTE detection approach proposed by \cite{Li_ICCV} cannot effectively detect perceptible backdoors in-flight. This approach blurs each test image (using, e.g., an average filter), hoping that if a perceptible backdoor pattern exists, it will be ``destroyed'' by the blurring filter, while most of the features corresponding to the source class still remain recognizable. Hence a detection could be made if the predicted label of a test image changes after blurring. While this defense is shown to be pretty effective against TTE attacks, where the image-specific perturbation for each test image is very subtle and could be easily destroyed, it cannot effectively defeat perceptible backdoor attacks in-flight, likely because blurring does not fully destroy the backdoor pattern and because the perceptible backdoor pattern has been robustly learned by the classifier during training.

\begin{table}[t]
	\begin{center}
		\caption{Percentage (\%) of predicted label changes for clean test images and backdoor test images when applying an average filter and a median filter to the images respectively, for each attack.}
		\resizebox{0.45\textwidth}{!}{
			\begin{tabular}{ |c|c|c|c|c| } 
				\hline
				& \multicolumn{2}{|c|}{avg. filter} & \multicolumn{2}{|c|}{med. filter}\\
				\hline
				& \thead{clean\\test image} & \thead{backdoor\\test image} & \thead{clean\\test image} & \thead{backdoor\\test image} \\ 
				\hline
				Attack A & 19.1 & 36.7 & 17.1 & 41.3\\
				\hline
				Attack B & 20.3 & 87.4 & 17.9 & 90.3\\
				\hline
				Attack C & 21.0 & 71.4 & 17.8 & 63.2\\
				\hline
				Attack D & 21.4 & 89.0 & 17.4 & 80.1\\
				\hline
				Attack E & 21.3 & 73.9 & 19.5 & 60.5\\
				\hline
				Attack F & 28.6 & 23.0 & 23.0 & 8.0\\
				\hline
				Attack G & 12.0 & 4.0 & 11.7 & 6.0\\
				\hline
				Attack H & 6.4 & 10.3 & 10.0 & 18.5\\
				\hline
				Attack I & 27.7 & 40.0 & 27.9 & 13.4\\
				\hline
			\end{tabular}\label{tab:TTE_in-flight}}
	\end{center}
\end{table}

In this experiment, we blur the clean test images (without considering whether they are correctly classified or not) and the test images with backdoors for each attack devised in Section \ref{sec:attack_crafting}, using an average filter and a median filter respectively. For Attack A--H, the filter size is $2\times2$, while for Attack I where the images have much higher resolution, the filter size is $10\times10$. In Table \ref{tab:TTE_in-flight}, we show the percentage of the clean test images and the percentage of the backdoor test images whose label prediction (by the classifier being attacked) is changed after blurring, when applying the average filter and the median filter respectively, for each attack. These two percentages are the false positive rate (FPR) and the true positive rate (TPR) of the detection approach. Only for Attack B and D, this blurring-based detector achieves a relative high TPR ($>80\%$), but at the cost of a non-negligible FPR ($>15\%$). For other attacks, further increasing the filter size may result in a higher TPR, but the FPR will increase correspondingly. Moreover, the filter size is a hyperparameter that cannot be easily determined in the in-flight defense scenario. Clearly, detecting perceptible backdoors in-flight is still an open problem -- in the absence of availability of the poisoned training set, post-training defense is currently the best defense solution to backdoor attacks for users of DNN classifiers.

\section{Conclusions and Future Work}\label{sec:conclusions}

In this paper, we proposed a detector for perceptible backdoors, post-training and {\it without access} to the (possible poisoned) training set. Our detector, inspired by two properties of perceptible backdoor patterns, is based on the maximum achievable misclassification fraction statistic. With an easily chosen threshold, our detector shows strong detection capability for a range of datasets and backdoor patterns.

Yet there are potential improvements to our detector that could be studied in the future. First, even for a DNN with relatively high test accuracy, there might be classes very similar to each other and with high class confusion; hence there might be a pattern found on a small spatial support that induces high group misclassification fraction for such a class pair. To avoid false detections, one may build knowledge of confusion matrix information into the detector, akin to what was done for imperceptible backdoor detection \cite{Post-TNNLS}. Second, the pattern estimation step in our detector is performed for all class pairs. When there are thousands of classes, a more efficient version of our detector should be of great interest.

\bibliographystyle{plain}
\bibliography{Biblio}

\end{document}